\pdfoutput=1
\documentclass{article} %
\PassOptionsToPackage{x11names}{xcolor}
\usepackage{iclr2026_conference,times}

\iclrpreprintcopy   

\usepackage{amsmath,amsfonts,bm}

\def\eqref#1{equation~\ref{#1}}

\def\1{\bm{1}}

\DeclareMathAlphabet{\mathsfit}{\encodingdefault}{\sfdefault}{m}{sl}
\SetMathAlphabet{\mathsfit}{bold}{\encodingdefault}{\sfdefault}{bx}{n}

\usepackage{hyperref}
\hypersetup{
    colorlinks=true,
    citecolor=RoyalBlue4,
    linkcolor=RoyalBlue4,
    urlcolor=RoyalBlue4
}
\usepackage{url}
\usepackage[capitalize,noabbrev]{cleveref}

\usepackage{booktabs} %
\usepackage{multirow}
\usepackage{xspace}
\usepackage{graphicx}
\usepackage[x11names]{xcolor}
\usepackage{caption}
\usepackage{textcomp}
\usepackage{subcaption}
\usepackage{enumitem}
\usepackage{wrapfig}
\usepackage{tcolorbox}
\tcbuselibrary{breakable}
\usepackage{listings}
\usepackage{tabularx}
\usepackage{array}
\usepackage{wrapfig}

\newcommand{\methodname}{$\textsc{MemIG}$\xspace}

\lstdefinelanguage{json}{
    basicstyle=\ttfamily\footnotesize,
    showstringspaces=false,
    breaklines=true,
    breakatwhitespace=false,
    postbreak={},
    literate=
        *{:}{{{\textcolor{teal}{:}}}}{1}
        {,}{{{\textcolor{gray}{,}}}}{1}
        {\{}{{{\textcolor{black}{\{}}}}{1}
        {\}}{{{\textcolor{black}{\}}}}}{1}
        {[}{{{\textcolor{black}{[}}}}{1}
        {]}{{{\textcolor{black}{]}}}}{1}
        {true}{{{\textcolor{blue}{true}}}}{4}
        {false}{{{\textcolor{red}{false}}}}{5}
        {null}{{{\textcolor{gray}{null}}}}{4},
    string=[s]{"}{"},
    stringstyle=\textcolor{black},
    comment=[l]{//},
    commentstyle=\textcolor{gray}\ttfamily,
    morestring=[b]',
    keywordstyle=\textcolor{teal},
}

\title{\methodname: Self-evolving Memory Construction \\ for Antibody Optimization}

\author{
    Yunhui Jang\textsuperscript{1}\thanks{Work done during an internship at Genentech $^\dag$ Correspondence} \quad
    Zach Wang\textsuperscript{1} \quad Jai Doshi\textsuperscript{2} \quad Kangway Chauang\textsuperscript{2}\\
    \ \textbf{Sidney Lisanza\textsuperscript{2}$^\dag$ \quad Taylor Joren \textsuperscript{2}$^\dag$}   \\
    \ \textsuperscript{1}KAIST, \textsuperscript{2} Genentech \\
    \ \texttt{yunhuijang@kaist.ac.kr}, \texttt{\{lisanza.sidney, joren.taylor\}@gene.com} \\
}

\definecolor{memigblue}{HTML}{3178E8}
\definecolor{memiggreen}{HTML}{00A651}
\definecolor{memigyellow}{HTML}{FFB800}

\begin{document}

\maketitle

\begin{abstract}
Memory has become central to self-evolving LLM agents, which store and retrieve experience from past rollouts to guide future decisions. Existing systems represent this experience as free-form natural language and maintain it by rewriting, which fits tasks whose experience is narratable. However, in scientific discovery and optimization, experience is often a quantity measured by an oracle under a specific context. Rewriting it as text leaves no index for where a record should be, replaces a measurement with a description, and cannot express how interventions interact. To address this, we argue that agent memory for biomolecule optimization should instead be indexable, quantitative, and structured, and we introduce \methodname. This memory construction framework realizes these three properties. In detail, \methodname addresses each insight with a domain-grounded identifier and accumulates the quantitative effect measured by a property oracle, so that updates remain local and retrieval becomes an exact lookup. Next, \methodname connects insights whose interventions are applied together and measures their epistasis by decomposing each evaluated set. Our results show that \methodname improves performance in most settings across in silico antibody optimization tasks.
\end{abstract}

\section{Introduction}

LLM agents can self-evolve by accumulating memories from previous rollouts, which guide future decisions~\citep{shinn2023reflexion, zhao2024expel, suzgun2026dynamic, zhang2026ace}. This offers a practical, economical alternative to fine-tuning, as the agent improves without expensive training while accumulated memory remains interpretable and editable. Nevertheless, the representation of this memory is free-form and unstructured, with experience from previous rollouts stored in natural language. This fits tasks whose experience is narratable, such as question answering and long-horizon conversation~\citep{zhong2024memorybank, packer2023memgpt, mem0}, where experience can be summarized in a sentence without losing rich information.

However, directly applying such narrative memory systems to biomolecule optimization is not straightforward. For instance, in a computational oracle-guided protein optimization task~\citep{notin2023proteingym, stanton2022lambo, gruver2023nos}, an agent proposes mutations to sequences, evaluates the candidate using oracles, and reuses measurements from the candidates to guide later proposals. Each evaluation is not merely a narratable experience, but a numerical measurement tied to a specific biological context, whose effect may depend on other co-applied mutations. Therefore, memory for biomolecule optimization should be \emph{indexable} by biological context, \emph{quantitative} in what it stores, and \emph{structured} to capture interactions between mutations. 

First, memory should be indexed by the context in which a measurement applies. Without such an index, there is no way to address a single entry, so incorporating a new experience requires an LLM to rewrite or summarize the entire memory. To address this, recent work~\citep{zhang2026ace, xu2025amem} introduces identifiers for each memory entry so that updates can be applied locally, but these identifiers are arbitrary labels without information about the context, so retrieval still relies on textual similarity. In contrast, the effect of the same mutation can depend on its exact biological context~\citep{notin2023proteingym, olsen2024germline}. Therefore, memory should address each entry by this context, so that updates and retrieval operate over the context rather than textual similarity.

Second, memory should preserve quantitative measurements rather than store them as free-form text. Existing memory methods store experience as natural language authored by an LLM~\citep{zhao2024expel, suzgun2026dynamic}, where a reported number is written rather than computed and is therefore not tied to the measurement that produced it. This can discard the quantitative signal required for scientific optimization. Memory should instead retain oracle outputs as persistent quantitative records, while the LLM only converts the retrieved records into interpretable guidance for the agent.

Third, memory should capture the measured dependence between entities, rather than semantic similarity between the textual records. Graph-based memories organize records as linked notes rather than a flat list, but their edges represent semantic relatedness between texts~\citep{mem0, xu2025amem}. However, in biomolecule optimization, the effect of one mutation can depend on the presence of another, a phenomenon known as epistasis. Memory should therefore record these interactions explicitly rather than assume that mutation effects are additive.

\begin{figure}
    \centering
    \includegraphics[width=0.98\linewidth]{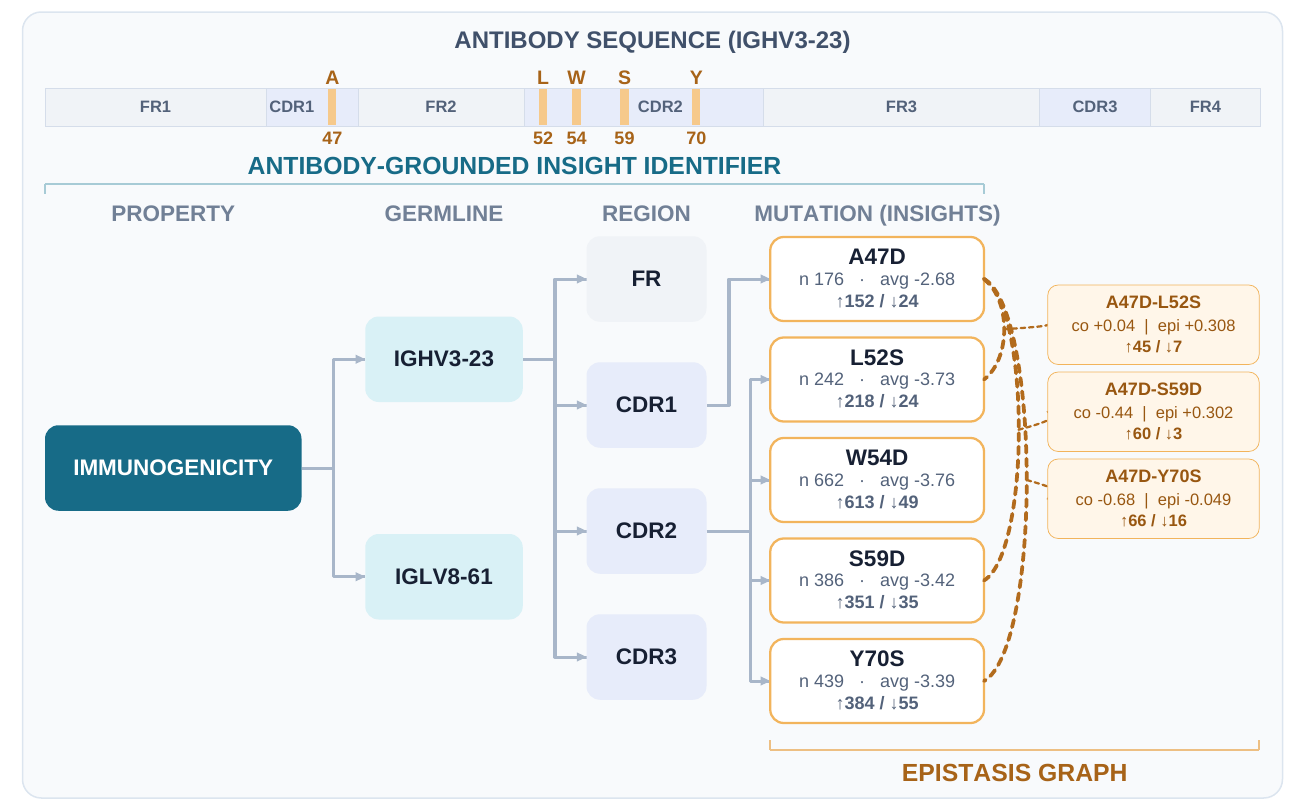}
    \caption{
        \textbf{Overview of \methodname.} \methodname constructs an epistasis graph with antibody-grounded identifiers. Directed edges organize property $\rightarrow$ germline $\rightarrow$ region $\rightarrow$ mutation hierarchies, while undirected edges connect co-occurring mutations and store their epistatic effects.
    }
    \label{fig:1_main}
    \vspace{-0.3in}
\end{figure}

To address this, we introduce \methodname (\textbf{Mem}ory as \textbf{I}ndexable \textbf{G}raphs), a domain-grounded memory construction framework for protein optimization agents, as described in ~\cref{fig:1_main}. We focus on the domain of antibody optimization and design. To make memory indexable, we define an antibody-grounded identifier that specifies the target property, the germline gene, the antibody region, and the AHo-numbered~\citep{honegger2001yet} mutation, so that each record is addressed by the biological context in which it applies. To make memory quantitative, each insight stores the effect computed by the oracle. 

Moreover, to build structured memory, \methodname creates a graph of co-occurring insights. Here, each edge in the graph connects mutations showing epistasis, quantified as the deviation of the joint effect from the sum of the individual effects. Next, before each rollout, \methodname retrieves insights from the graph based on the context identifier, ranks them by their measured effects, and provides the selected insights with their graph neighborhoods to the agent. Then an LLM renders them as interpretable guidance while the quantitative memory remains. We empirically showed that incorporating \methodname into frontier LLMs~\citep{glm5team2026glm5vibecodingagentic, Anthropic2025Haiku45} improves performance in the majority of settings in in silico antibody optimization tasks.

Our contributions are summarized as follows:
\begin{itemize}[leftmargin=*, itemsep=2pt, topsep=2pt]
    \item We introduce \methodname, an \emph{indexable}, \emph{quantitative}, and \emph{structured} memory system which addresses each insight with an antibody-grounded identifier and stores the effect computed by a property oracle, so that updates remain local and retrieval becomes an exact lookup.
    \item We construct a graph-based memory over insights, where an edge stores the epistatic effect of mutations, so that memory precisely records how mutations combine.
    \item We demonstrate that MemIG outperforms existing memory systems on antibody optimization.
\end{itemize}

\section{Method: \methodname}

\begin{wrapfigure}{r}{0.55\linewidth}
    \centering
    \vspace{-0.2in}
    \includegraphics[width=\linewidth]{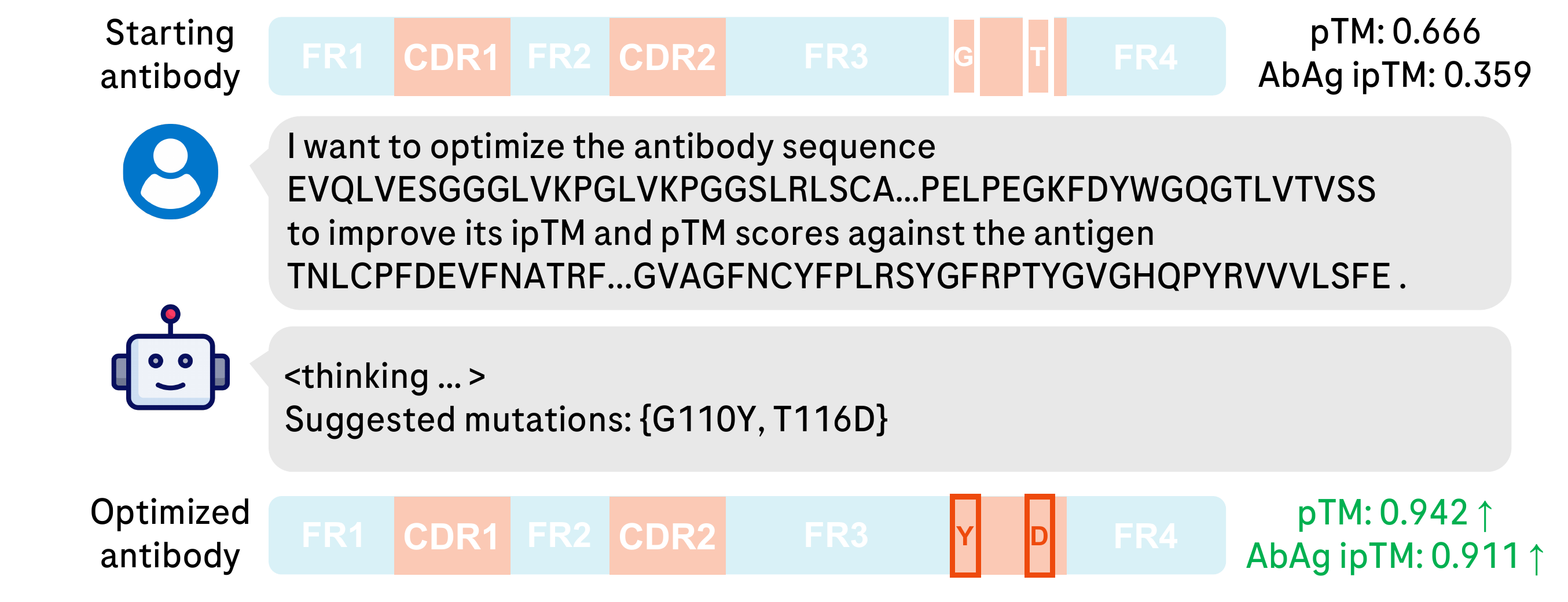}
    \caption{\textbf{Antibody optimization task.}}
    \vspace{-0.1in}
    \label{fig:3_task_description}
    \vspace{-0.1in}
\end{wrapfigure}

In this section, we present \methodname (\textbf{Mem}ory as \textbf{I}ndexable \textbf{G}raphs), an indexable, quantitative, and structured memory system for antibody optimization agents. 
In this work, we consider the following formulation of antibody optimization: given an antibody sequence, the model proposes a set of mutations to improve a defined set of properties under a maximum edit distance, as described in \cref{fig:3_task_description}.

Throughout this section, an \textit{observation} is a single measured effect of applying mutations, and an \textit{insight} is the summary of the observations sharing the same identifier. \textit{Memory} is then the graph whose nodes are these insights and whose edges store how mutations interact when applied together, as illustrated in \cref{fig:1_main}. Memory is self-evolving as the agent writes observations from its own evaluations, constructs memory from the observations, and retrieves the relevant insights.

Each component of \methodname realizes three properties of memory: indexability, quantitativeness, and structured dependencies. \cref{sec:2_1_antibody_identifier} defines the indexable identifiers and quantitative records, \cref{sec:2_2_epistasis} describes the epistasis graph, and \cref{sec:2_3_retrieval} explains retrieval based on the constructed epistasis memory graph.

\subsection{Antibody-grounded insight identifier}\label{sec:2_1_antibody_identifier}

\paragraph{Indexable identifier.} The identifier makes memory indexable by addressing each insight with the antibody context where it was observed. In detail, an identifier is composed of a property, germline, region and mutation, where the region is a contiguous block of residues that define a framework or complementarity-determining region (e.g. FR1 or CDR2) and the mutation position comes from AHo alignment ~\citep{honegger2001yet}, such as \texttt{immunogenicity/IGHV3-23/[cdr1]/H:A47D}. The AHo scheme enables a single insight to accumulate evidence across antibodies of different lengths. We use exact matching so that observations are aggregated only when their biological context and substitution match. Consequently, the same mutation observed under a different context forms a different insight. This design is inspired by \citet{zhang2026ace}, which introduces identifiers to enable local memory updates.

Addressing insights in this way keeps the memory update local and efficient. Each new observation updates only the insight sharing its identifier, leaving the rest of memory untouched. Therefore, the cost scales with the number of identifiers affected by the new observation rather than the total memory size, whereas rewriting the entire memory grows more expensive as experience accumulates.

\paragraph{Quantitative record.} Each insight records the quantitative effect of its mutation, so that memory reports exact quantities rather than free-form impressions. In detail, an insight accumulates the mean effect across turns on each property computed by the oracle, the number of observations, and the directional votes over every observation sharing the identifier. The directional vote records whether the mutation moved the property toward the objective, so that the insight reports not only the average magnitude of an effect but also how consistently that effect occurs. For example, the insight
\begin{center}
\texttt{immunogenicity/IGHV3-23/H:A47D [cdr1]: avg -2.68, n=176, $\uparrow$152/$\downarrow$24}
\end{center}
records the substitution of alanine (A) with aspartate (D) at heavy-chain position 47 in the CDR1 of a sequence from the IGHV3-23 gene family. The mutation decreased (improved) immunogenicity by 2.68 on average over 176 observations, and did so in all but 24 of them.

\subsection{Epistasis graph}\label{sec:2_2_epistasis}

The epistasis graph makes memory structured by connecting insights whose mutations interact. The joint effects of mutations may be non-additive to other co-occurring mutations, a phenomenon known as epistasis, so a set of independent insights cannot represent it. To capture this in the epistasis graph, each node is a single mutation, and each edge accumulates the joint effect of a co-occurring pair. These edges connect mutations within a chain and across the heavy and light chains.

We measure the epistatic effect of a pair by decomposing each mutation set the agent proposes. For each pair of mutations $a$ and $b$ in the set, we additionally apply $a$ alone, $b$ alone, and $a$ and $b$ together to the starting sequence, and measure the resulting property changes $\Delta_{a}$, $\Delta_{b}$, and $\Delta_{ab}$. Each node stores the mean isolated effect when the mutation is applied alone as \texttt{single\_avg}, while \texttt{avg} stores the mean effect of the complete evaluated mutation sets containing that mutation to reflect the effects of co-applied mutations.

\begin{figure}
    \centering
    \includegraphics[width=0.98\linewidth]{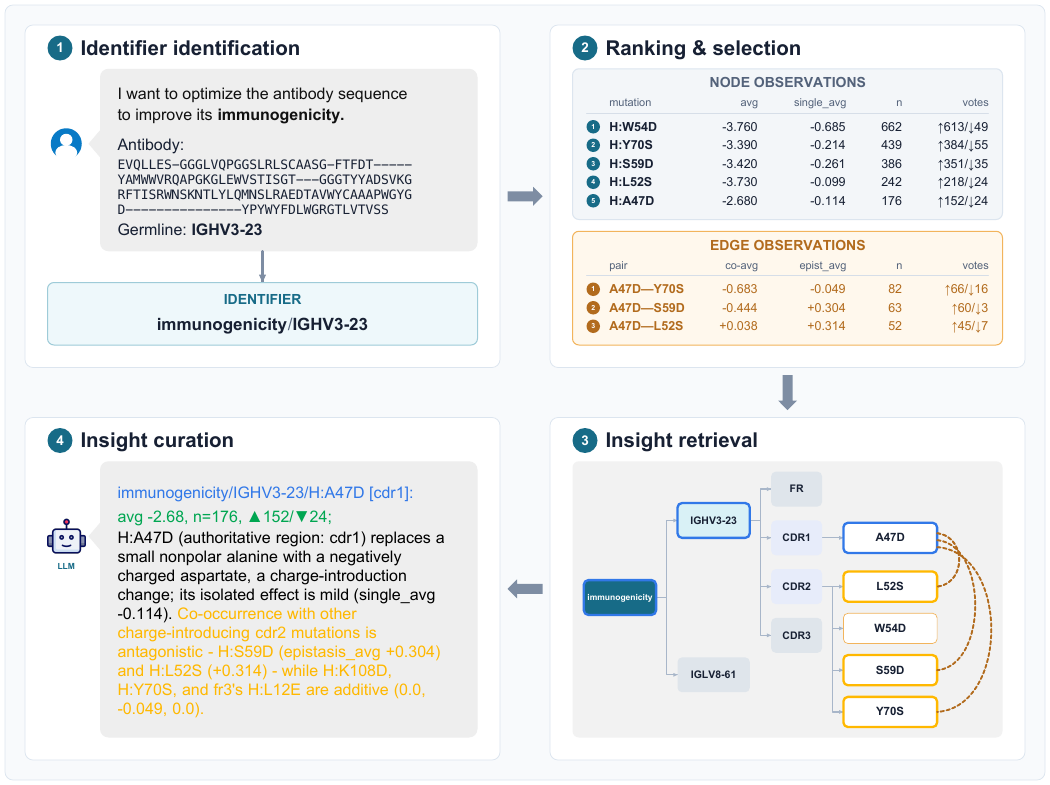}
    \caption{
        \textbf{Memory retrieval of \methodname.} Given a starting antibody, \methodname identifies the matching identifiers, retrieves the corresponding insights with their neighbors, and curates them into natural language. \textcolor{memigblue}{Blue} denotes the antibody-grounded identifier and the corresponding node, \textcolor{memiggreen}{green} its quantitative record, and \textcolor{memigyellow}{yellow} the epistasis graph neighbors and the curation based on them. 
    }
    \label{fig:3_3_memory_example}
    \vspace{-0.2in}
\end{figure}

The epistatic effect is then the deviation of the pair from additivity, $\varepsilon_{ab} = \Delta_{ab} - \Delta_{a} - \Delta_{b}$. When $\varepsilon_{ab}=0$, the joint effect equals the sum of the individual effects, indicating additivity and no epistasis. A nonzero $\varepsilon_{ab}$ indicates an epistatic interaction: positive values mean that the joint property change exceeds the sum of the individual changes, whereas negative values mean that it falls below this sum. The edge between $a$ and $b$ stores the mean of $\Delta_{ab}$ as \texttt{co\_avg} and the mean of $\varepsilon_{ab}$ as \texttt{epistasis\_avg} over every observation of that pair. Notably, this decomposition requires additional oracle calls, which is possible because property oracles are cheap and re-runnable on any sequence. An example of the stored nodes and edges with these values is shown in \cref{fig:3_3_memory_example}.

\subsection{Memory retrieval}\label{sec:2_3_retrieval}

Based on the constructed graph, memory is retrieved before each rollout and injected into the prompt. The retrieval process consists of four steps: identifier identification, ranking and selection, insight retrieval, and insight curation, as described in \cref{fig:3_3_memory_example}. In detail, we collect the insights whose identifiers match the germline and the target properties of the current sequence,  and whose mutations match its wild-type residues at the corresponding AHo
positions. We rank them by their net directional votes breaking ties by the magnitude of the \texttt{avg}, and select up to $K$ insights. In addition to these exact-match insights, the memory also includes global insights that aggregate mutation trends across germlines to capture broader mutation trends, as detailed in \cref{appx:2_experiments}.

The retrieved insights are curated into natural language so that the agent receives interpretable guidance. Specifically, we curate each insight with an LLM, which also receives up to $N$ 1-hop neighbor insights and their edges from the epistasis graph, so that the text describes both the isolated and joint effects. Curation produces the guidance presented to the agent without modifying the stored quantitative
statistics. For example, an insight \texttt{immunogenicity/IGHV3-23/H:A47D [cdr1]} is curated as:

\begin{center}
\texttt{
immunogenicity/IGHV3-23/H:A47D [cdr1]: avg -2.68, n=176, $\uparrow$152/$\downarrow$24; 
H:A47D (authoritative region: cdr1) replaces a small nonpolar alanine with a negatively charged aspartate, a charge-introduction change; its isolated effect is mild (single\_avg -0.114), so the -2.68 average (n=176) is mostly confounded by companion mutations. Co-occurrence with other charge-introducing cdr2 mutations is antagonistic — H:S59D (epistasis\_avg +0.304) and H:L52S (+0.314) — while H:K108D, H:Y70S, and fr3's H:L12E are additive (0.0, -0.049, 0.0). 
}
\end{center}

After curation, the curated insights are concatenated to form the memory prompt. Memory is initialized as empty and the memory of the sequence is written only after all of its rollouts have been scored. Therefore, the rollouts of a sequence share the same memory snapshot, and a sequence retrieves only the insights accumulated from the sequences optimized before it. We provide an example of the constructed memory in \cref{appx:1_examples}.

\section{Experiments}\label{sec:3_experiment}

In this section, we evaluate \methodname's effectiveness on two antibody optimization tasks. We employ Haiku-4.5~\citep{Anthropic2025Haiku45} and GLM-5.3-Flash~\citep{glm5team2026glm5vibecodingagentic} as our backbone models. We provide the detailed experimental settings in \cref{appx:2_experiments}.

\paragraph{Metrics.} We employ two metrics throughout the experiments: pass@$k$~\citep{chen2021evaluating} and improvement (Imp.). Here, we consider two types of objectives: (1) minimization and maximization, which move the property as far as possible in one direction, and (2) range, which moves the property into a target interval. Pass@$k$ estimates the probability that at least one of $k$ sampled rollouts passes, computed as $1 - \binom{n-c}{k}/\binom{n}{k}$, where $n$ rollouts are sampled and $c$ pass. We report $k \in \{1, 3\}$. A rollout passes when the optimized sequence improves upon the starting sequence for minimization and maximization tasks, or lies within the target interval for range tasks.

Next, improvement measures the magnitude of the gain. Specifically, it is computed as the signed difference between the optimized and the starting property values in the favorable direction for minimization and maximization tasks, and as the reduction in distance to the target interval for range tasks. For example, in the range task, when the starting sequence property is $-0.6$, and the optimized sequence property is $-0.4$, the distance to the target interval $[-0.5, -0.2]$ decreases from $0.1$ to $0.0$, giving an improvement of $0.1$.  Notably, both metrics are computed at the best turn of each rollout determined by the reward and are then averaged over all rollouts.

\paragraph{Baselines.} We utilize six baselines, including two statistical and four LLM-based approaches. The statistical baselines include random and MCMC~\citep{1953JChPh21.1087M}. The random baseline samples $k$ mutations uniformly from $\{1, \dots, B\}$ mutations, where $B$ is the edit distance budget, and applies the mutations to the starting variable fragment (Fv). The MCMC baseline runs Metropolis-Hastings from the starting Fv. At each of $T$ steps, matched to the turn budget of the agent, it proposes three random mutations that remain within the edit distance budget, scores them with the same reward $r$ used to evaluate the agent, and accepts the mutations under the Metropolis criterion $\min(1, \exp(\Delta r / \tau))$ with the temperature fixed at $\tau = 1.0$. 

The LLM-based baselines include a vanilla LLM agent and three LLM agents with different memory frameworks, including observations-only, DC-Cu~\citep{suzgun2026dynamic}, and ACE~\citep{zhang2026ace}. The vanilla LLM agent applies the agentic system without memory. Next, the LLM agent with observations receives the same oracle outputs that \methodname uses to construct its memory, but in their raw form rather than aggregated under identifiers or curated. Lastly, DC-Cu and ACE are self-evolving memory baselines that accumulate experience across rollouts, where DC-Cu regenerates a compact cheatsheet after each query while ACE maintains an indexed playbook updated through incremental deltas. For fair comparison, we provide all LLM-based baselines with the same tool access, prompt, maximum input and output tokens, and turn budget, which isolates the contribution of the memory.

\subsection{Multiple property optimization}\label{sec:3_2_multi}

\paragraph{Task description.} The goal of multiple property optimization is to satisfy multiple property constraints at once by mutating the variable fragment (Fv) of a given antibody sequence. Specifically, we select four antibody properties: immunogenicity~\citep{pep2vec}, liability count~\citep{biopython}, GRAVY~\citep{gravy}, and charge. Here, immunogenicity quantifies the risk that the antibody triggers an unwanted immune response and liability count measures chemically unstable sequence motifs (e.g., oxidation and deamidation sites). Next, GRAVY measures the average hydrophobicity over the residues and charge measures the net charge of the sequence at pH 5.5. Immunogenicity and liability count are minimization, and GRAVY and charge are range tasks, targeting $[-0.5, -0.2]$ and  $[0.0, 8.0]$, respectively. We provide the details in \cref{appx:2_experiments}.

We randomly select 100 antibody sequences from the OAS dataset~\citep{oas} and generate 10 rollouts per sequence. We filter out the sequences that already satisfy the target constraint, so the pass@$k$ of the starting sequences is zero by construction.

\begin{figure}
    \centering
    \includegraphics[width=\linewidth]{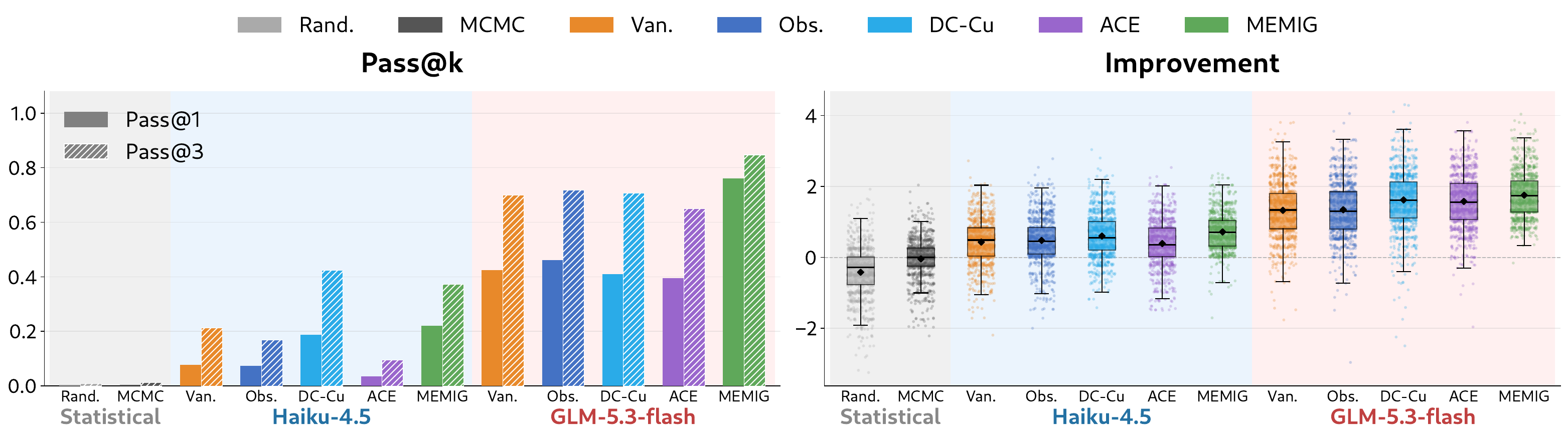}
    \caption{
        \textbf{Multi property optimization results.} Pass@1 and pass@3 (left) and the distribution of improvement over the starting sequence (right), where each colored dot is one rollout and the black diamond marks the mean.
    }
    \label{fig:4_multi_prop}
    \vspace{-0.2in}
\end{figure}

\paragraph{Results.} We provide the results in \cref{fig:4_multi_prop} and full values in \cref{appx:3_add_experiments}. We observe that \methodname consistently improves multiple property optimization across both backbones. Specifically, it achieves the highest average improvement on Haiku-4.5 (0.718 vs.\ 0.602 for the strongest baseline) and GLM-5.3-Flash (1.754 vs.\ 1.621), while also attaining the best pass@1 on both backbones. The box plots show a broad upward shift in the improvement, with medians close to the corresponding means, indicating that the gains are systematic rather than driven by a few favorable rollouts. Across all baseline comparisons, these gains are significant under one-sided Mann--Whitney U tests~\citep{mann1947test} with Holm--Bonferroni correction ($p < 0.005$).

\subsection{In silico antibody binding optimization}\label{sec:3_3_insilico}

\paragraph{Task description.} The goal of this task is to optimize the binding of an antibody to a given antigen by mutating CDR-H3. Specifically, we optimize two confidence scores of a structure prediction model, the predicted TM score (pTM) for the antibody structure and the interface predicted TM score (AbAg-ipTM) for the antibody-antigen interaction, where we target $>0.8$ and $>0.7$ respectively. Such confidence scores discriminate binders from non-binders and therefore serve as useful proxies for in silico binding optimization~\citep{abramson2024accurate, lin2022language, Zhang2026.04.10.717613}. We employ Protenix-v2~\citep{Zhang2026.04.10.717613} for its predictive power and open availability.

We select antibody-antigen complexes from the SAbDab ~\citep{sabdab}. We retain only complexes that are (i) of type fragment antigen-binding (Fab), (ii) bound to a single antigen chain, and (iii) whose CDR-H3 loop makes substantial contact with the antigen interface (H3 interface fraction $>0.3$). These selected complexes are grouped into $11$ clusters, where each cluster is treated as one optimization task, generating 30 rollouts per complex. We provide the details includng the number of complexes and their PDB IDs in each cluster in \cref{appx:2_experiments}.

\begin{figure}
    \centering
    \includegraphics[width=\linewidth]{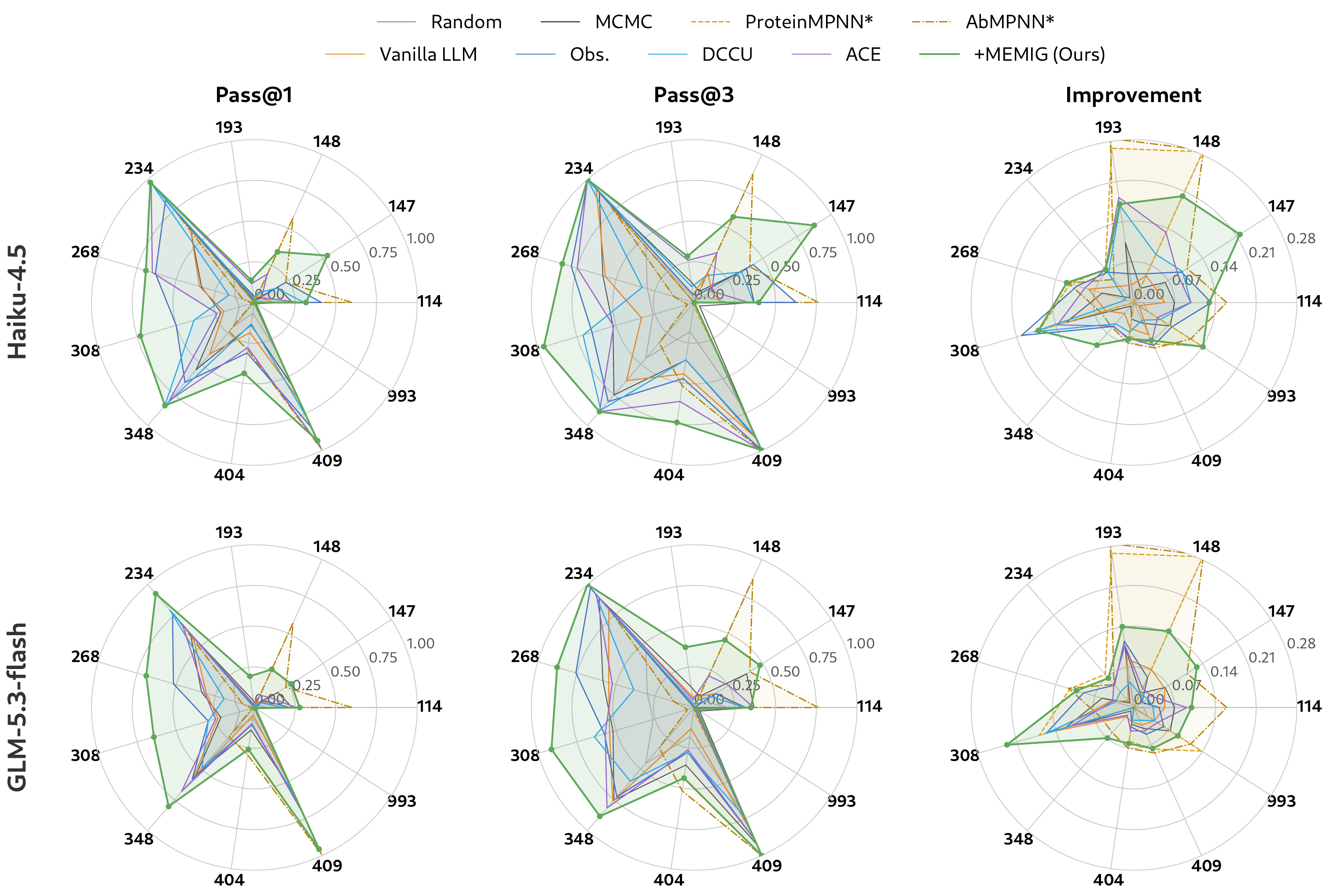}
    \vspace{-0.2in}
    \caption{
        \textbf{In silico antibody binding optimization.} Each numeric label corresponds to an antibody-antigen cluster, clustered by CDR sequence identity. $^{*}$ denotes the structural baselines (dashed), which have access to the complex structure.
    }
    \label{fig:3_1_denovo}
    \vspace{-0.2in}
\end{figure}

\paragraph{Baselines.} We add two structural baselines for in silico antibody optimization task, including ProteinMPNN~\citep{dauparas2022robust} and AbMPNN~\citep{dreyer2023inverse}. These baselines inverse-fold the native complex with every position fixed except CDR-H3. It is notable that they receive the backbone coordinates of the antibody-antigen complex, whereas \methodname observes the sequence alone. We sample the same number of designs as the agent's turn budget and evaluate all designs with the same oracle. We treat these methods as structure-informed references because they receive complex backbone coordinates, whereas MEMIG receives only sequences. They could also be incorporated into the agent as design tools.

\paragraph{Results.} We provide the results in \cref{fig:3_1_denovo} and the detailed values in \cref{appx:3_add_experiments}. Among the non-structural methods, we observe that \methodname attains the best pass@1 in 9 of the 11 clusters with GLM-5.3-Flash and in 7 of the 11 clusters with Haiku-4.5, and the highest improvement in every cluster with GLM- 5.3-Flash. Compared to all LLM-based and statistical baselines within each backbone, these gains are statistically significant under one-sided Mann--Whitney U tests~\citep{mann1947test} with Holm--Bonferroni correction ($p < 0.005$). The structural baselines included as reference anchors lead on improvement, since improvement is averaged over all designs regardless of whether they pass the constraint. Notably, ProteinMPNN attains a pass rate of zero in 8 of the 11 clusters while leading on improvement in three of them, which shows that a large improvement does not imply that the designs reach the target.

\subsection{Analysis}

We conduct the following analyses on the in silico antibody binding optimization task with GLM-5.3-Flash on three randomly picked clusters (148, 193, 993). We provide additional analyses including same oracle call analysis and details in \cref{appx:3_add_experiments}.

\begin{wraptable}{r}{0.45\linewidth}
\vspace{-0.1in}
\centering
\caption{\textbf{Ablation study on component}.}
\label{tab:3_4_ablation_comp}
\small
\begin{tabular}{@{}c ccc@{}}
\toprule[1.25pt]
\textbf{Component} & Pass@1 & Pass@3 & Imp. \\
\midrule[1.25pt]
{\methodname (Ours)} & 0.149 & 0.278 & 0.125 \\
\midrule
{Identifier} & 0.074 & 0.153 & 0.117 \\
\midrule
Epistasis & 0.049 & 0.130 & 0.142 \\
\midrule
{Vanilla LLM} & 0.015 & 0.047 & 0.067 \\
\bottomrule[1.25pt]
\end{tabular}
\end{wraptable}

\paragraph{Ablation study on component.} We ablate the components of \methodname. In detail, the indexable identifier drops the germline+region field which keeps property and mutation only. Next, the epistasis graph removes the edges, so that memory stores the information of each mutation without access to the joint effects. As shown in \cref{tab:3_4_ablation_comp}, we observe that every ablation degrades the pass rate, which shows that all components contribute to satisfying the constraints. In contrast, the ablated epistasis attain a higher improvement, since improvement is averaged over all designs regardless of whether they satisfy the constraint.

\paragraph{Memory faithfulness.} We next verify that the agent actually uses the retrieved insights stored in memory by \methodname. In detail, we take the insights that \methodname retrieves for each starting antibody as a fixed reference set, and compare MEMIG’s proposals with this reference set. The adoption measures the fraction of proposed mutations that appear anywhere in the reference set, while helpful and harmful adoption further split this fraction by whether the recorded effect of the matched insight points toward or against the objective based on their directional votes. Since a mutation has a separate insight for each property, it can be helpful for one and harmful for another, so the two fractions sum to more than the total adoption. Additionally, we report the net directional alignment, a gap between helpful and harmful adoption ($\Delta$), as the agent must not only read the memory but also distinguish its direction.

\begin{wraptable}{r}{0.48\textwidth}
\vspace{-\intextsep}
\centering
\caption{\textbf{Memory faithfulness.} \methodname raises both adoption and net directional alignment.}\label{tab3_4_2:faithfulness}
\small
\setlength{\tabcolsep}{5pt}
\begin{tabular}{cccc}
\toprule[1.25pt]
 Adoption & Helpful & Harmful & $\Delta$ \\
\midrule[1.25pt]
 0.931 & 0.909 & 0.004 & 0.905 \\
\bottomrule[1.25pt]
\end{tabular}
\vspace{-\intextsep}
\end{wraptable}

We provide the results in \cref{tab3_4_2:faithfulness}. We observe that 93.1\% of the proposals by \methodname were present in the memory. These mutations were mostly from helpful insights, not from harmful insights, which shows that the agent reads and understands the recorded direction of an insight.

\begin{figure}
    \centering
    \includegraphics[width=\linewidth]{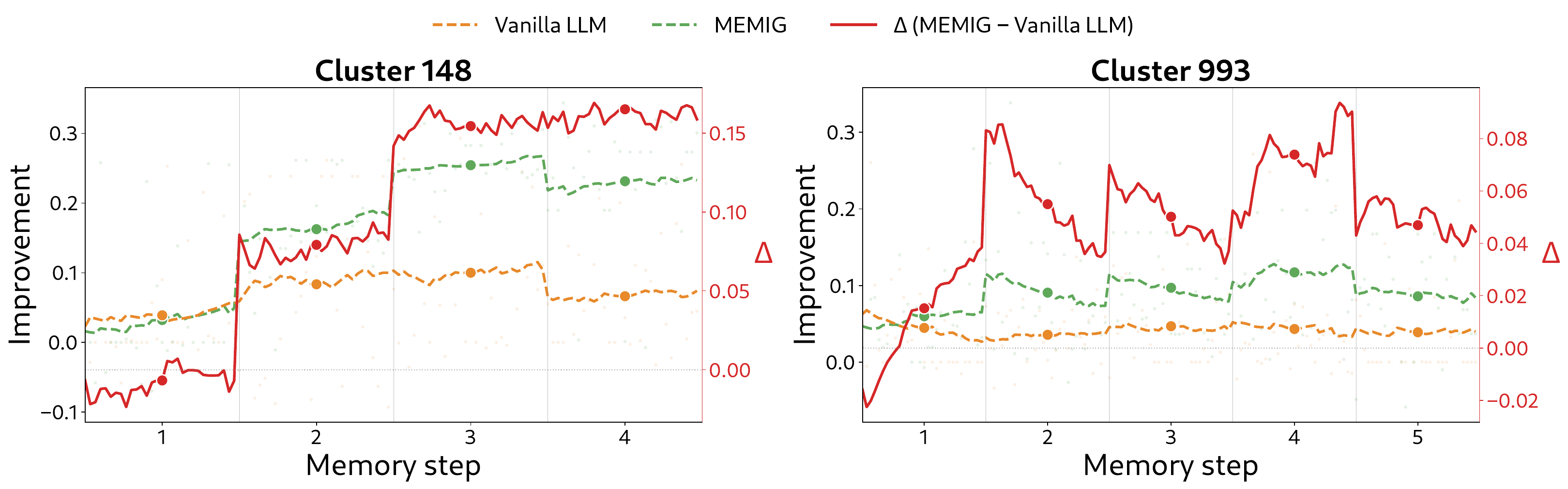} 
    \caption{
        \textbf{Cumulative performance progression.} Vertical lines mark memory updates, where the rollouts between two lines are generated concurrently under the same memory. The curves report the centered rolling mean with a half-width of 30 rollouts, computed within each memory step. Notably, the ordering of rollouts within the same memory is random, so a downward slope does not imply performance decrease. Each dot marks the mean over all rollouts of that memory step.
    }
    \label{fig:3_6_cumulative}
    \vspace{-0.2in}
\end{figure}

\paragraph{Cumulative performance progression.} We analyze how performance changes as memory self-evolves in \cref{fig:3_6_cumulative} by tracking the improvement of the vanilla LLM and \methodname at each memory step, where the rollouts sharing a memory step are generated concurrently. We report clusters 148 and 993, whose germline remains constant across the starting complexes, so that the insights accumulated from earlier rollouts apply directly to later ones. We observe that \methodname improves over memory versions while the vanilla LLM stays near flat, so the gain comes from the accumulated memory rather than from the agent itself. Notably, the gap widens after the first memory update and remains positive, which shows that \methodname improves through self-evolution. We additionally report the same analysis without the rolling average in \cref{appx:3_add_experiments}.

\section{Related work}

\paragraph{Self-evolving agent with memory.} Self-evolving agents have recently gained substantial attention, where agents accumulate experience from past rollouts and store it into external memory without updating model weights~\citep{shinn2023reflexion}. ExpeL~\citep{zhao2024expel} abstracts natural language insights from successful and failed experience, and maintains them through LLM-judged upvotes and downvotes. Dynamic Cheatsheet~\citep{suzgun2026dynamic} generates the solution based on the current cheat sheet, curates the insight based on the generation, and updates the cheat sheet, while ACE~\citep{zhang2026ace} instead maintains an itemized playbook with helpful and harmful counters, updated with incremental deltas.

Graph-based memories keep their entries as linked notes rather than a flat list, where nodes and edges represent semantic relations between textual memories~\citep{mem0, xu2025amem}. Nevertheless, these approaches store experience as natural language written and judged by an LLM, address entries by opaque labels that do not encode applicability context, and connect entries by textual relatedness. To address these challenges, \methodname stores oracle-measured quantities under the biological context-derived identifiers and connects them through oracle-measured interactions.

\paragraph{LLM agents for protein optimization.} Protein sequence optimization has been studied as black-box optimization over a discrete space, where a model is trained on previously evaluated sequences and is used to propose the next candidates~\citep{stanton2022lambo, gruver2023nos, notin2023proteingym}. Recently, LLM agents have entered this pipeline by orchestrating property predictors as tools and reasoning over the returned scores to select or revise candidates, with some accumulating design strategies across rollouts as natural language~\citep{roohani2025biodiscoveryagent, tan2026rankandreasonmultiagentcollaborationaccelerates, liu2026expertise}. However, these approaches only accumulate narrative feedback without any quantitative record, which is not indexed and not structured in terms of biological context. To address this, \methodname keeps oracle-measured effects under antibody-grounded insight identifiers and connects them through measured pairwise interactions, which keeps accumulated memory indexable, quantitative, and structured across rollouts.

\section{Conclusion}

In this paper, we introduce \methodname, a memory construction framework for antibody optimization agents that keeps accumulated observations indexable, quantitative, and structured rather than narrated. Each insight is addressed by an antibody-grounded identifier over the property, germline, region, and mutation, and stores the effect measured by a property oracle. Next, edges between insights record the epistatic effect based on how mutations behave when applied together. Our findings suggest that the unit of memory should be defined by the domain it serves, and that grounding memory in measured quantities rather than generated text allows experience to accumulate without summarization. Although we instantiate MEMIG for antibody optimization, the framework can extend to other oracle-guided optimization problems when the domain provides a suitable identifier. 

\paragraph{Limitation and challenge.} First, our evaluation is conducted in silico using computational property oracles. This bounds the cost model of the epistasis decomposition, which assumes oracles that can be re-run freely. Although these oracles provide a controlled testbed, improvements may not directly translate to experimentally measured properties. Second, \methodname currently represents mutation interactions through pairwise epistasis edges. Although each edge is estimated by evaluating the individual mutations and their pair, multi-mutation designs may exhibit higher-order interactions.  Next, the curation step of \methodname relies on the capability of the backbone model, as in prior work~\citep{zhang2026ace, suzgun2026dynamic}. If the model fails to describe a retrieved insight faithfully, the curated guidance may be misleading while the quantitative record mitigates this.

\newpage

\section*{Ethics statement}

This work studies memory construction for antibody optimization agents, and the experiments are conducted in silico with computational property oracles. We use only publicly released data, including antibody sequences from OAS~\citep{oas} and antibody-antigen complexes from SAbDab~\citep{sabdab}, under their respective licenses; the OAS sequences derive from published immune repertoire studies and are not linked to identifiable individuals, and no human subjects are involved in our experiments. \methodname optimizes developability and predicted interface confidence rather than any property related to toxicity or pathogenicity, and any resulting candidate would require experimental validation and human review before use. 

\section*{Reproducibility statement}

We provide the information required to reproduce our experiments in the main text and appendix. \cref{sec:3_experiment} describes the evaluation setup and baselines, while \cref{appx:2_experiments} provides the property oracles, datasets, prompts, and hyperparameters. Complete numerical results and additional analyses are reported in \cref{appx:3_add_experiments}.

\section*{AI Disclosure}
We used generative AI tools to polish writing of the manuscript, create figures, and assist with analysis and plotting code. The research hypothesis, the method design, and all experiments were determined by the authors. The remaining tasks requiring disclosure, such as generating synthetic datasets and writing proofs, are not applicable.

\newpage
\bibliographystyle{iclr2026_conference}
\bibliography{iclr2026_conference}

\newpage
\appendix

\setcounter{table}{0}
\renewcommand{\thetable}{A\arabic{table}}

\section{Examples}\label{appx:1_examples}

\subsection{Example reasoning trace}

\begin{tcolorbox}[
    colback=gray!5!white,
    colframe=gray!60!teal,
    title={
        \parbox[t]{\dimexpr\linewidth-4mm\relax}{%
            \ttfamily Reasoning trace
        }
    },
    breakable
]

\textbf{Turn 1 $|$ Tool result (baseline)}
\begin{lstlisting}[language=json]
{
  "gravy": -0.1851, "charge_ph5_5": -0.154,
  "sequence_liabilities": 7, "immunogenicity": 5.0,
  "edit_distance": 0, "sequence_length": 228
}
\end{lstlisting}

\textbf{Turn 2 $|$ Assistant}\\
Baseline: GRAVY $-$0.1851 (need $\leq -0.2$), charge $-$0.154 (need $\geq 0$), liabilities 7, immunogenicity 5.0, length 228 \checkmark. Need to lower gravy slightly and raise charge at pH 5.5. Prioritizing charge-raising mutations with big effects and gravy-lowering, drawing on \textbf{MEMORY} insights for IGHV3-23: H:V13K and H:D83K give the largest charge gains ($+$5.1, $+$5.26), while H:M41T, H:F78S, and H:M93L are top gravy-reducers with fully additive co-occurrence (epistasis\_avg $=$ 0.0 across all tested partners).

\begin{lstlisting}[language=json]
tool_call: evaluate_mutations
{
  "heavy_mutations": "{V13K, L12K, M41T, F78S, M93L, D83K, S7K, D100N, T88A}",
  "light_mutations": "{}"
}
\end{lstlisting}

\textbf{Turn 2 $|$ Tool result}
\begin{lstlisting}[language=json]
{
  "gravy": -0.2776, "charge_ph5_5": 5.778,
  "sequence_liabilities": 5, "immunogenicity": 7.0,
  "edit_distance": 9
}
\end{lstlisting}

\textbf{Turn 3 $|$ Assistant}\\
Gravy and charge are now in range, and liabilities dropped from 7 to 5, but immunogenicity rose from 5.0 to 7.0 --- worse than baseline. This contradicts the \textbf{global insight} that ``to\_K'' mutations are broadly disfavored for immunogenicity ($-$2.66 avg globally); apparently the four lysine substitutions (V13K, L12K, D83K, S7K) compound this penalty in combination even though their individual single-mutation deltas measured 0.0. Trimming to fewer K mutations and testing a leaner, more additive panel.

\begin{lstlisting}[language=json]
tool_call: evaluate_mutations
{
  "heavy_mutations": "{M41T, F78S, D83K, D100N}",
  "light_mutations": "{}"
}
\end{lstlisting}

\textbf{Turn 3 $|$ Tool result}
\begin{lstlisting}[language=json]
{
  "gravy": -0.214, "charge_ph5_5": 2.778,
  "sequence_liabilities": 6, "immunogenicity": 4.0,
  "single_mutations": [
    {"token": "D100N", "immunogenicity_delta": -1.0},
    {"token": "M41T", "sequence_liabilities_delta": -1}
  ]
}
\end{lstlisting}

\textbf{Turn 4 $|$ Assistant}\\
This 4-edit set improves every objective: gravy $-$0.214 \checkmark, charge 2.778 \checkmark, liabilities 6 (from 7), immunogenicity 4.0 (from 5.0). Per \textbf{MEMORY}, D100N is the only mutation with a measured $-$1.0 direct immunogenicity effect, and it composes additively with D83K (epistasis\_avg $\approx$ 0). Testing whether a light-chain liability fix can push further without disturbing this balance --- the light FR1 has a methionine (M4) analogous to the heavy H:M41 oxidation-liability site the memory repeatedly flags.

\begin{lstlisting}[language=json]
tool_call: evaluate_mutations
{
  "heavy_mutations": "{M41T, F78S, D83K, D100N}",
  "light_mutations": "{M4L}"
}
\end{lstlisting}

\textbf{Turn 4 $|$ Tool result}
\begin{lstlisting}[language=json]
{
  "gravy": -0.2057, "charge_ph5_5": 2.778,
  "sequence_liabilities": 5, "immunogenicity": 4.0,
  "single_mutations": [
    {"token": "M4L", "chain": "L", "sequence_liabilities_delta": -1}
  ]
}
\end{lstlisting}

\textbf{Turn 5 $|$ Assistant}\\
The methionine-removal analogy paid off: L:M4L drops liabilities to 5 with zero cost to gravy, charge, or immunogenicity, all pairwise epistases measuring $\approx$0. Final design: heavy \{M41T, F78S, D83K, D100N\}, light \{M4L\} --- all four objectives improved or in range, 5 of 10 edits used.

\end{tcolorbox}

\subsection{Observations}

\begin{tcolorbox}[
    colback=gray!5!white,
    colframe=gray!60!teal,
    title={
        \parbox[t]{\dimexpr\linewidth-4mm\relax}{%
            \ttfamily Observations
        }
    },
    breakable
]
\begin{lstlisting}[language=json]
{
  "applied_mutations": [
    {"bullet_id": "H:M41L", "chain": "H", "germline": "IGHV3-21*08",
     "group_id": "H:41L", "pos": 41, "region": "cdr1", "target": "L",
     "target_group": "nonpolar", "token": "M41L", "wt": "M", "wt_group": "nonpolar"},
    {"bullet_id": "H:N75T", "chain": "H", "germline": "IGHV3-21*08",
     "group_id": "H:75T", "pos": 75, "region": "cdr2", "target": "T",
     "target_group": "polar", "token": "N75T", "wt": "N", "wt_group": "polar"},
    {"bullet_id": "H:E92K", "chain": "H", "germline": "IGHV3-21*08",
     "group_id": "H:92K", "pos": 92, "region": "fr3", "target": "K",
     "target_group": "positive", "token": "E92K", "wt": "E", "wt_group": "negative"},
    {"bullet_id": "H:F31S", "chain": "H", "germline": "IGHV3-21*08",
     "group_id": "H:31S", "pos": 31, "region": "cdr1", "target": "S",
     "target_group": "polar", "token": "F31S", "wt": "F", "wt_group": "aromatic"},
    {"bullet_id": "H:L91T", "chain": "H", "germline": "IGHV3-21*08",
     "group_id": "H:91T", "pos": 91, "region": "fr3", "target": "T",
     "target_group": "polar", "token": "L91T", "wt": "L", "wt_group": "nonpolar"},
    {"bullet_id": "L:L54S", "chain": "L", "germline": "IGLV1-44*01",
     "group_id": "L:54S", "pos": 54, "region": "cdr2", "target": "S",
     "target_group": "polar", "token": "L54S", "wt": "L", "wt_group": "nonpolar"},
    {"bullet_id": "L:Y57S", "chain": "L", "germline": "IGLV1-44*01",
     "group_id": "L:57S", "pos": 57, "region": "cdr2", "target": "S",
     "target_group": "polar", "token": "Y57S", "wt": "Y", "wt_group": "aromatic"},
    {"bullet_id": "L:D99N", "chain": "L", "germline": "IGLV1-44*01",
     "group_id": "L:99N", "pos": 99, "region": "fr3", "target": "N",
     "target_group": "polar", "token": "D99N", "wt": "D", "wt_group": "negative"},
    {"bullet_id": "L:F78S", "chain": "L", "germline": "IGLV1-44*01",
     "group_id": "L:78S", "pos": 78, "region": "cdr2", "target": "S",
     "target_group": "polar", "token": "F78S", "wt": "F", "wt_group": "aromatic"},
    {"bullet_id": "L:W109Y", "chain": "L", "germline": "IGLV1-44*01",
     "group_id": "L:109Y", "pos": 109, "region": "cdr3", "target": "Y",
     "target_group": "aromatic", "token": "W109Y", "wt": "W", "wt_group": "aromatic"}
  ],
  "example_id": 1,
  "germline_heavy": "IGHV3-21*08",
  "germline_light": "IGLV1-44*01",
  "memory_version": 1,
  "n_edits": 10,
  "passed": 1.0,
  "properties": {
    "charge_ph5_5": {"beneficial": true, "delta": 2.884, "improvement": 1.3, "start": -1.3, "value": 1.584},
    "gravy": {"beneficial": true, "delta": -0.052, "improvement": 0.0092, "start": -0.1908, "value": -0.2428},
    "immunogenicity": {"beneficial": true, "delta": -4.0, "improvement": 4.0, "start": 18.0, "value": 14.0},
    "sequence_liabilities": {"beneficial": true, "delta": -2.0, "improvement": 2.0, "start": 7.0, "value": 5.0}
  },
  "pair_mutations (sample)": [
    {"token_a": "H:M41L", "token_b": "H:N75T",
     "immunogenicity": {"beneficial": true, "delta": -1.0, "epistasis": 0.0},
     "sequence_liabilities": {"beneficial": true, "delta": -2.0, "epistasis": 0.0}},
    {"token_a": "H:E92K", "token_b": "H:L91T",
     "immunogenicity": {"beneficial": true, "delta": -1.0, "epistasis": 1.0},
     "charge_ph5_5": {"beneficial": false, "delta": 1.918, "epistasis": 0.0}},
    {"token_a": "L:D99N", "token_b": "L:F78S",
     "immunogenicity": {"beneficial": true, "delta": -1.0, "epistasis": -1.0},
     "gravy": {"beneficial": true, "delta": -0.0158, "epistasis": 0.0}}
  ],
  "proposed_seq": "--QLVES-GGGLVQPGGSLRLSCVASG-FTSSP-----FCLTWVRQAPGKGLEWISSISYF--DATYIYYTHSLTGRITVSRDNAQNSLFTKINSLRAEDTAVYYCARSRGIG-------------------NYYFDLWGQGTLVTVSS|QPVLTQP-PSASEAPGQGVRISCSGGI-SNIGT-----TAVQWYRQIPGAAPQSLISE--------NNQRPSGVPDRSSGSRSG--TSASLDISRLQSNDEADYYCATYDDS---------------------LQGYVFGTGTKVTVL",
  "reward": 0.9810355955754664,
  "rollout_idx": 0,
  "run": "bfa94dd6",
  "start_seq": "--QLVES-GGGLVQPGGSLRLSCVASG-FTFSP-----FCMTWVRQAPGKGLEWISSISYF--DATYIYYTHSLNGRITVSRDNAQNSLFLEINSLRAEDTAVYYCARSRGIG-------------------NYYFDLWGQGTLVTVSS|QPVLTQP-PSASEAPGQGVRISCSGGI-SNIGT-----TAVQWYRQIPGAAPQLLIYE--------NNQRPSGVPDRFSGSRSG--TSASLDISRLQSDDEADYYCATWDDS---------------------LQGYVFGTGTKVTVL",
  "turn_idx": 0
}
\end{lstlisting}
\end{tcolorbox}

\subsection{Epistasis graph}

\begin{tcolorbox}[
    colback=gray!5!white,
    colframe=gray!60!teal,
    title={
        \parbox[t]{\dimexpr\linewidth-4mm\relax}{%
            \ttfamily Epistasis graph
        }
    },
    breakable
]
\begin{lstlisting}[language=json]
{
  "property": "immunogenicity",
  "germline_heavy": "IGHV3-23",
  "germline_light": "IGKV3-15",

  "nodes (H:cdr1)": {
    "H:25T": {
      "avg": -3.83, "support": 6, "upvotes": 6, "downvotes": 0,
      "group_note": "H:25T lowers immunogenicity on average (-3.83, n=6).",
      "bullets": {
        "H:A25T": {
          "avg": -3.83, "support": 6, "single_avg": 0.0,
          "insight": "H:A25T lowered immunogenicity by -3.83 on average (n=6)."
        }
      }
    },
    "H:41L": {
      "avg": -2.02, "support": 194, "upvotes": 158, "downvotes": 36,
      "group_note": "H:41L lowers immunogenicity on average (-2.02, n=194).",
      "bullets": {
        "H:M41L": {
          "avg": -2.02, "support": 194, "single_avg": 0.043,
          "insight": "H:M41L lowered immunogenicity by -2.02 on average (n=194)."
        }
      }
    },
    "H:43F": {
      "avg": 3.54, "support": 13, "upvotes": 2, "downvotes": 11,
      "group_note": "H:43F raises immunogenicity on average (+3.54, n=13).",
      "bullets": {
        "H:W43F": {
          "avg": 3.54, "support": 13, "single_avg": 1.25,
          "insight": "H:W43F raised immunogenicity by +3.54 on average (n=13)."
        }
      }
    },
    "H:45D": {
      "avg": -3.0, "support": 3, "upvotes": 3, "downvotes": 0,
      "group_note": "H:45D lowers immunogenicity on average (-3, n=3).",
      "bullets": {
        "H:R45D": {
          "avg": -3.0, "support": 3, "single_avg": 0.0,
          "insight": "H:R45D lowered immunogenicity by -3 on average (n=3)."
        }
      }
    }
  },

  "edges (intra-chain, H-H)": {
    "H:F90K__H:N87T": {
      "co_avg": -3.132, "co_support": 106, "co_up": 106, "co_down": 0,
      "epistasis_avg": 1.009,
      "note": "Large-support pair; positive epistasis means the combo underperforms the sum of single benefits."
    },
    "H:F90K__H:S88K": {
      "co_avg": -3.206, "co_support": 107, "co_up": 107, "co_down": 0,
      "epistasis_avg": 0.617,
      "note": "High-confidence pair (n=107), mild positive epistasis but still strongly beneficial."
    },
    "H:D83K__H:F90K": {
      "co_avg": -3.0, "co_support": 8, "co_up": 8, "co_down": 0,
      "epistasis_avg": 0.0,
      "note": "Additive charge-reversal + aromatic-to-K pairing."
    },
    "H:I58S__H:M93L": {
      "co_avg": 0.632, "co_support": 57, "co_up": 36, "co_down": 21,
      "epistasis_avg": 0.316,
      "note": "Net-harmful pair with synergistic (positive) epistasis; avoid combining."
    }
  },

  "edges (cross-chain, H-L)": {
    "H:F78S__L:I92T": {
      "co_avg": -3.0, "co_support": 24, "co_up": 24, "co_down": 0,
      "epistasis_avg": 0.0,
      "note": "Strong beneficial cross-chain pair, fully additive."
    },
    "H:D83K__L:M4L": {
      "co_avg": 0.0, "co_support": 36, "co_up": 36, "co_down": 0,
      "epistasis_avg": 0.0,
      "note": "Neutral, high-support cross-chain pairing (FR3 charge reversal + light N-term liability fix)."
    },
    "H:M41T__L:M4T": {
      "co_avg": 0.173, "co_support": 52, "co_up": 43, "co_down": 9,
      "epistasis_avg": 0.0,
      "note": "Largest-support cross-chain pair in the set; net mildly harmful but additive."
    },
    "H:R97D__L:M4T": {
      "co_avg": -3.0, "co_support": 14, "co_up": 14, "co_down": 0,
      "epistasis_avg": 0.0,
      "note": "Strong, additive cross-chain deimmunizing pair."
    }
  }
}
\end{lstlisting}
\end{tcolorbox}

\subsection{Rendered memory}

  \begin{tcolorbox}[
    colback=gray!5!white, 
    colframe=gray!60!teal,
    title={
      \parbox[t]{\dimexpr\linewidth-4mm\relax}{%
        \ttfamily
        Rendered memory
      }
    },
    breakable
  ]

example\_id: 98 \\
germline\_heavy: IGHV3-23*01 \\
germline\_light: IGKV1-8*01 \\
props: gravy, charge\_ph5\_5, sequence\_liabilities, immunogenicity \\

\# MEMORY \\
Prior rollout insights. Use them to guide mutation choices, but still verify with tools. \\
Node single\_avg = isolated single-mutation effect. Edge co\_avg = isolated pair effect (only those two mutations). epistasis\_avg = co\_avg - (single\_avg\_A + single\_avg\_B), the deviation from additivity in the raw property: positive means co\_avg is greater than the sum of the singles, negative means less, near zero is additive. Favorability depends on the objective direction (raise / lower / target band), not the raw sign of the deviation. Section grouping reflects the objective direction; $\uparrow$/$\downarrow$ votes encode favorability. \\

\textbf{[LOCAL BENEFICIAL INSIGHTS (sequence\_liabilities/immunogenicity)]}

Mutations whose average effect helps the objective 

(avg = mean property change; $\uparrow$/$\downarrow$ = rollouts it helped/hurt). 

Prefer these.
\begin{itemize}
\item sequence\_liabilities/IGHV3-23/H:M41T [cdr1]: avg $-$3.09, n=1196, $\uparrow$1195/$\downarrow$1; In the IGHV3-23 cdr1, H:M41T replaces a nonpolar methionine with polar threonine, removing a methionine oxidation liability and lowering sequence\_liabilities by $-$3.09 on average (n=1196). CDR residues can contact antigen, so swapping hydrophobic bulk for a small polar side chain should be checked against paratope geometry. Co-occurrence data show purely additive behavior with H:F78S, H:M93L, H:V13K, H:L12K, and H:D83K (all epistasis\_avg $=$ 0.0), so M41T can be combined freely with these liability-reducing mutations.

\item immunogenicity/IGHV3-23/H:F78S [cdr2]: avg $-$1.7, n=894, $\uparrow$772/$\downarrow$122; H:F78S (cdr2) swaps an aromatic residue for a polar one, eliminating a bulky hydrophobic side chain that likely contributes T-cell receptor/MHC-facing hydrophobic patches, and yields the largest immunogenicity reduction in this set (deterministic effect $-$1.70, n=894). Because F78 sits in CDR2, verify antigen binding after the aromatic-to-polar swap, since removing the F side chain could perturb paratope packing. Combinations are favorable: synergistic with H:I80T (epistasis\_avg $+$0.200, co\_avg $-$0.438 versus single\_avg $-$0.274) and with H:M93L (epistasis\_avg $+$0.069), while V13K, L12K, S7K, and M41T are additive (epistasis\_avg 0.0), supporting F78S as a stackable core of a deimmunization design.
\dots 
\end{itemize}

\vspace{\baselineskip}
\textbf{[LOCAL INCREASING INSIGHTS (charge\_ph5\_5)]} \\
Mutations that raise the property on average (avg = mean change; $\uparrow$/$\downarrow$ = times it helped/hurt reaching the target band --- e.g. if already in/above the band, raising it usually hurts). Use when you need it higher.
\begin{itemize}
\item charge\_ph5\_5/IGHV3-23/H:D83K [fr3]: avg $+$5.26, n=698, $\uparrow$697/$\downarrow$1; H:D83K in fr3 reverses a negative Asp to a positive Lys, removing a $-$1 group and adding a $+$1 group at pH 5.5, making it the strongest and cleanest charge lever here ($+$5.26, n=698, just 1 downvote). Its pair with H:D109K in cdr3 is additive --- co\_avg $=$ 3.93 versus D109K single\_avg $=$ 1.97, epistasis\_avg $=$ $+$0.0002 --- and other pairs (H:M114L, H:V89T, H:N111S, H:M113L, H:M41L) all show epistasis\_avg $=$ 0.0, so Asp-to-Lys reversals stack without antagonism. Favor D-to-K reversals for fr3 charge engineering, but check that the eliminated acidic side chain does not participate in a needed salt bridge before committing.

\item charge\_ph5\_5/IGHV3-23/H:V13K [fr1]: avg $+$5.1, n=764, $\uparrow$761/$\downarrow$3; H:V13K in fr1 replaces a nonpolar Val with a positively charged Lys, which is directly protonated at pH 5.5 and delivers the largest charge\_ph5\_5 gain in this set ($+$5.1, n=764, only 3 downvotes) --- a mechanistically consistent direct-charge mechanism. Pairing with H:D100A in fr3 is additive: co\_avg $=$ 1.97 versus D100A's single\_avg $=$ 0.97, epistasis\_avg $\approx$ 0, and all other neighbors also show epistasis\_avg $=$ 0.0. Use V13K as a first-line fr1 lever for raising acidic-pH charge, stacking it with other charge mutations without expecting antagonism.
\dots
\end{itemize}

\vspace{\baselineskip}
\textbf{[LOCAL DECREASING INSIGHTS (gravy)]} \\
Mutations that lower the property on average ($\uparrow$/$\downarrow$ = times it helped/hurt reaching the target band --- e.g. if already in/below the band, lowering it usually hurts). Use when you need it lower.
\begin{itemize}
\item gravy/IGHV3-23/H:L91K [fr3]: avg $-$0.133, n=273, $\uparrow$272/$\downarrow$1; H:L91K in fr3 substitutes the nonpolar leucine with a positively charged lysine, reliably lowering hydrophobicity (single\_avg $-$0.133, n=273, 272 upvotes), making it a strong lever for reducing overall GRAVY when the framework can tolerate added positive charge. All six measured pairings are essentially additive (epistasis\_avg within $\pm$4e$-$5), e.g. with H:M41T in cdr1 (epistasis $+$1.7e$-$5) and H:D83K in fr3 ($-$4.1e$-$5), so its effect should transfer predictably into multi-mutation designs.

\item gravy/IGHV3-23/H:M41T [cdr1]: avg $-$0.102, n=1196, $\uparrow$1181/$\downarrow$15; H:M41T (cdr1) replaces a nonpolar methionine with polar threonine, lowering gravy by $-$0.102 on average (n=1196) --- a strong shift toward surface hydrophilicity that typically aids solubility when CDR1 is solvent-exposed, but may hurt if M41 packs against a hydrophobic binding interface. The effect is essentially additive with co-mutations: epistasis with H:F78S is $-$7.6e$-$06, with H:I80T 2.1e$-$05, and with H:V13K $-$1.2e$-$05, so no pair-level risk of antagonism.
\dots
\end{itemize}

\vspace{\baselineskip}
\textbf{[GLOBAL BENEFICIAL INSIGHTS (sequence\_liabilities/immunogenicity)]}

Mutations whose average effect helps the objective (avg = mean property change; $\uparrow$/$\downarrow$ = rollouts it helped/hurt). Prefer these.

\begin{itemize}
\item global/sequence\_liabilities/to\_K: Substituting to K has a helpful average effect on sequence liabilities ($-$2.85), with 6244 of 6272 aggregated observations agreeing on help. Because this rule aggregates across all germlines with strongly consistent votes, it is likely generalizable rather than germline-specific. Apply this rule broadly as a default avoid-mutation for sequence liabilities, and defer to a germline-specific local insight only if that germline's local data contradict the global direction.
\item global/immunogenicity/to\_T: Substituting to T has a helpful average effect on immunogenicity ($-$3.31), with 2247 of 2456 aggregated observations voting helpful. The large negative average and consistent votes across all germlines indicate a generalizable, germline-independent penalty, and it is the largest immunogenicity help among these targets. Apply this rule broadly by default; defer only when a germline-specific local insight shows the opposite trend in that germline.
\dots
\end{itemize}

\vspace{\baselineskip}
\textbf{[GLOBAL INCREASING INSIGHTS (charge\_ph5\_5)]}

Mutations that raise the property on average (avg = mean change; $\uparrow$/$\downarrow$ = times it helped/hurt reaching the target band). Use when you need it higher.

\begin{itemize}
\item global/charge\_ph5\_5/to\_K: Mutation to lysine raises charge at pH 5.5 strongly and consistently, averaging $+$5.81 across 6272 observations with a $\sim$99.5\% upvote rate, consistent with lysine's protonation and positive charge at acidic pH. This effect reflects the direct physicochemical consequence of adding a protonatable amine, so it is expected to generalize across germlines with little germline-specific variance. Apply this rule broadly whenever the goal is increasing charge at pH 5.5.
\dots
\end{itemize}

\vspace{\baselineskip}
\textbf{[GLOBAL DECREASING INSIGHTS (gravy/charge\_ph5\_5)]}

Mutations that lower the property on average ($\uparrow$/$\downarrow$ = times it helped/hurt reaching the target band). Use when you need it lower.

\begin{itemize}
\item global/gravy/to\_K: Mutating to K lowers GRAVY (hydropathy) by $\sim-$0.135 on average, a harmful shift for hydrophobicity with 6272 observations and 99\% agreement. The huge support base makes this direction highly generalizable across germlines, since K's charged side chain consistently reduces hydropathy regardless of framework context. Apply this rule broadly as a default expectation; defer to germline-specific insights mainly when magnitude, not direction, matters.
\item global/charge\_ph5\_5/to\_D: Mutating to D reduces charge at pH 5.5 by an average of $-$1.13, a large negative charge shift. The effect is nearly universal: 389 of 398 instances (98\%) agree, with only 9 counterexamples, reflecting the consistent electrostatic consequence of introducing an acidic residue. Treat this as a broadly applicable rule across germlines; deviate only when a local insight documents an exception.
\dots
\end{itemize}

\end{tcolorbox}

\newpage
\section{Experimental details}\label{appx:2_experiments}

\subsection{Details of optimized properties}

We provide the details of the optimized properties, including the oracle and the description of their significance in \cref{tab:3_2_1_mpo_descriptions}.

\begin{table*}[h]
    \centering
    \caption{\textbf{Optimized properties.} Six antibody properties the agent optimizes, the oracle or function measuring each, and why each are significant.}\label{tab:3_2_1_mpo_descriptions}
    \small
    \begin{tabularx}{\linewidth}{l|l|l|>{\raggedright\arraybackslash}X}
    \toprule[1.25pt]
        \textbf{Property} & \textbf{Objective} & \textbf{Oracle} & \textbf{Why it matters} \\
        \midrule[1.25pt]
        Immunogenicity & Minimize & Pep2Vec ~\citep{pep2vec} & Predicted MHC class~II epitope cores that can provoke an anti-drug antibody response. \\
        Liabilities & Minimize & Biopython ~\citep{biopython} & Motifs that degrade during manufacture and storage, costing potency and shelf life. \\
        GRAVY & Range & Biopython ~\citep{gravy} & Exposed hydrophobicity drives aggregation, but an overly hydrophilic domain loses stability. \\
        Charge & Range & Biopython ~\citep{biopython} & Endosomal charge at pH~5.5 governs clearance rate and formulation viscosity at both extremes. \\
        pTM & Range & Protenix~\citep{Zhang2026.04.10.717613} & Predicted TM-score between the predicted and the true complex. \\
        AbAg iPTM & Range & Protenix~\citep{Zhang2026.04.10.717613} & Accuracy of the predicted relative placement of the antibody and the antigen. \\
     \bottomrule[1.25pt]
    \end{tabularx}
\end{table*}

\subsection{Details of in silico binding optimization dataset}

We provide the details of in silico binding optimization dataset in \cref{tab:denovo_clusters}.Notably, an example is identified by a complex identifier of the form \texttt{\{index\}\_\{pdb\}\_\{antibody chains\}\_\{antigen chain\}}, so two entries from the same PDB with different chain combinations are two distinct examples.

\begin{table}[h]
    \centering
    \caption{\textbf{Details of the in silico binding optimization dataset.} Antibodies were clustered by CDR sequence identity. Each antibody's Heavy and Light chain CDR loops are concatenated into one string. Mmseqs2~\citep{steinegger2017mmseqs2} is run with a 70\% sequence identity cutoff.}\label{tab:denovo_clusters}
    \vspace{0.1in}
    \begin{tabular}{ccc}
    \toprule[1.25pt]
    \textbf{Cluster} & \textbf{\# Complexes} & \textbf{PDBs} \\
    \midrule[1.25pt]
    993  & 5  & 7l7e, 8d8q, 8ekd, 8suo \\
    114 & 6  & 6kyz, 8f2j \\
    147 & 7  & 7ben, 7beo, 7orb, 7tbf \\
    148 & 4  & 7jmw, 7lq7, 7s5q, 7s5r \\
    193 & 6  & 4fp8, 5umn, 6d0u \\
    234 & 4  & 4mwf, 6bkb, 6uyd, 6uym \\
    \multirow{2}{*}{268} & \multirow{2}{*}{24} & 4g6f, 4u6g, 5ghw, 5iq7, 5iq9, 5sy8, 5t29, 5t5b, 5t6l, 5t80, 5t85, \\
               &    & 5tfw, 5wdf, 6vpx, 6x58, 8sx3, 8tzw, 8u08, 8v2e \\
    308 & 7  & 6p3r, 6p3s \\
    348 & 8  & 7won, 7pqz, 7swp, 7wp0, 7wp1, 7wp2, 7wp5, 8siq \\
    404 & 5  & 6vi1, 6xmi, 8djg \\
    409 & 3  & 2hrp, 8gkl, 8v8f \\
    \bottomrule[1.25pt]
    \end{tabular}
    
\end{table}

\subsection{Details of global insight construction}

Global insights capture mutation trends that are shared across germlines in addition to the context-specific insights retrieved by exact matching. In detail, we aggregate observations for each property by their mutation target, such as \texttt{to K}, and record the mean property effect, number of observations, and directional votes. These insights are stored separately from the exact-match insights and are included in the memory prompt together with the retrieved local insights. Global insights provide broader evidence when a mutation trend transfers across biological contexts. Notably, they complement rather than replace the context-specific insights defined by the antibody-grounded identifier. 

\subsection{Prompts}

  \begin{tcolorbox}[
    colback=gray!5!white, 
    colframe=gray!60!black,
    title={
      \parbox[t]{\dimexpr\linewidth-4mm\relax}{%
        \ttfamily
        Shared prompt for multi property and in silico  antibody binding optimization
      }
    },
    breakable
  ]
  
Your task is to optimize an antibody Fv by proposing AHO-coordinate mutations \
to its heavy and light variable domains, while staying within a maximum number \
of de-gapped sequence edits from the starting molecule.

\vspace{\baselineskip}

TURN BUDGET

You have \{max\_turns\} turns total. Each turn is either a mutation-set proposal OR a tool call — both count toward the limit. A mutation-set proposal does NOT end the episode. After every proposal, you must call evaluate\_mutations in a separate message to evaluate it, then use the measured properties to propose an improved mutation set. Propose a different mutation set on every non-tool-call turn.

\vspace{\baselineskip}

COORDINATES

Positions use AHO (Honegger-Plueckthun) antibody numbering: a fixed coordinate system in which a given column maps to the same structural site across antibodies. A position means the canonical AHO column, not the raw de-gapped offset. A "-" is a canonical AHO position that is absent in this antibody.

Count columns INCLUDING the "-" gaps in the displayed sequence — every "-" still consumes a column number. To place a token, read the residue printed at that column in the chain shown below and use it as the token's first letter; if that letter is not what you intended to edit, your column index is wrong (you have almost certainly counted the de-gapped offset instead of the AHO column). \\

Example: for ''...G A - S G..." at columns "...24 25 26 27 28...", the S sits at column 27, so editing it is S27X — not S26X, because the gap at column 26 still counts.

\vspace{\baselineskip}

OUTPUT FORMAT

Each proposal must contain all three tag blocks:

\textless\textbar reasoning\_summary\_start\textbar \textgreater

[Your reasoning: what mutations you made, why, how you expect the target property to change]

\textless\textbar reasoning\_summary\_end\textbar \textgreater

\textless\textbar mutation\_answer\_heavy\_start\textbar \textgreater

\{{A27D, G31S, T58A}\}

\textless\textbar mutation\_answer\_heavy\_end\textbar \textgreater

\textless\textbar mutation\_answer\_light\_start\textbar \textgreater

\{{S27T, Y50F}\}

\textless \textbar mutation\_answer\_light\_end\textbar \textgreater

Mutation token syntax on AHO columns: \\
- A27D changes AHO column 27 from A to D (a substitution — the common case). \\
- -50G fills an absent AHO column 50 with G (insertion). \\
- A95- empties AHO column 95, deleting that residue from the de-gapped sequence. \\
- \{{}\} makes no change. \\

Most edits are simple substitutions; reach for the gap operations only when an \
insertion or deletion is actually needed.

Before proposing edits, call \
get\_antibody\_residue\_index(heavy\_reference, light\_reference) once. Use the \
returned \{{wt}\}\{{pos}\} prefixes verbatim and append only your target residue \
(`A27` -\textgreater `A27D`) so your wildtype letter and AHO column are always correct.

Positions never renumber because AHO chain length is fixed. Your edit budget is \
{max\_edits} de-gapped edits across both chains combined: residue-to-residue \
substitutions count as 1, residue-to-gap counts as 1, and gap-to-residue counts \
as 1. Gap-to-gap changes do nothing and should not be proposed.

\vspace{\baselineskip}

VALID MUTATION RULES

1. Match the wild-type letter. The first character of every token MUST be the residue currently at that AHO column in the chain shown below, using "-" if the  column is a gap. A mismatch is skipped. \\
2. Stay in range. Heavy positions must be between 1 and \{heavy\_aho\_length\}; light positions must be between 1 and \{light\_aho\_length\}. Out-of-range tokens are skipped. \\
3. No duplicates. Do not target the same AHO column twice within one chain; duplicate-position tokens are skipped. \\

evaluate\_mutations reports any mutation it had to skip and why, such as expected V at column 27. When you see this, re-read the displayed chain at that exact column, set your token's first letter to the residue printed there, and resubmit — do not keep the rejected token. A skipped mutation is one that simply did not happen.\\

When a separate MEMORY message is present, it contains prior rollout insights for related germlines and objectives; use it to prioritize mutations, but still verify every proposal with evaluate\_mutations.\\

To evaluate a mutation set, first propose it in one message, then call evaluate\_mutations in a separate message, passing your mutation set and the original starting chains as heavy\_reference and light\_reference. It returns both the reconstructed sequence and its properties.

\vspace{\baselineskip}

ITERATION RULE

For range objectives, keep proposing and evaluating until the target value is inside the requested range and all constraints hold, or until you exhaust the turn budget. For maximize/minimize objectives, keep improving and evaluating until the turn budget is exhausted.

\vspace{\baselineskip}

AVAILABLE PROPERTIES

\{available\_properties\}

\vspace{\baselineskip}

REQUIREMENTS

1. Mutation-set proposals and tool calls must be in separate messages. \\
2. Your FIRST message must be a tool-call message that calls get\_antibody\_residue\_index \
and measures the starting Fv baseline. \\
3. The reasoning summary must be substantive. \\
4. The heavy and light mutation answer tags must contain ONLY edit sets — no labels. \\
5. Your LAST message must include both heavy and light mutation answer tags. \\

GOAL \\
Given the antibody Fv AHO-numbered heavy and light chains below, propose \
designs within \{max\_edits\} combined de-gapped edits that \{objective\_text\}. \\

SEQUENCE REGION MAP \\
Heavy chain (AHO-numbered, \{heavy\_aho\_length\} columns): \\
  Full Fv:        \{heavy\_aho\_sequence\}  \\
  FR1:            \{h\_fr1\}  \\
  CDR1:           \{h\_cdr1\}  \\
  FR2:            \{h\_fr2\}  \\
  CDR2:           \{h\_cdr2\}  \\
  FR3:            \{h\_fr3\}  \\
  CDR3:           \{h\_cdr3\}  \\
  FR4:            \{h\_fr4\}  \\
  Germline:       \{h\_v\_gene\} / \{h\_j\_gene\} (\{h\_species\})  \\

Light chain (AHO-numbered, \{light\_aho\_length\} columns):  \\
  Full Fv:        \{light\_aho\_sequence\}  \\
  FR1:            \{l\_fr1\}  \\
  CDR1:           \{l\_cdr1\}  \\
  FR2:            \{l\_fr2\}  \\
  CDR2:           \{l\_cdr2\}  \\
  FR3:            \{l\_fr3\}  \\
  CDR3:           \{l\_cdr3\}  \\
  FR4:            \{l\_fr4\}  \\
  Germline:       \{l\_v\_gene\} / \{l\_j\_gene\} (\{l\_species\})  \\

Objective:
\{objective\_detail\}

Constraints:
\{constraints\_text\} \\

WORKFLOW

1. In the first tool-call message, call get\_antibody\_residue\_index on the starting AHO chains and call evaluate\_mutations with \{{}\} for heavy\_mutations and \{{}\} for light\_mutations to measure the baseline.  \\
2. Identify heavy and/or light AHO-column substitutions likely to move the target property. \\
3. Propose heavy and light mutation sets in a separate message. \\
4. Call evaluate\_mutations on your heavy and light mutation sets plus the starting AHO chains to evaluate them. \\
5. Iterate — evaluate every proposal, then propose an improved mutation set. For range objectives, stop early only after the value is in range and constraints hold. For maximize/minimize objectives, keep improving until you run out of turns. \\
6. Your last message must contain both mutation answer tag blocks with your best edit sets. \\

  \end{tcolorbox}

\paragraph{Property optimization.} The task prompt is combined with the shared prompt.

  \begin{tcolorbox}[
    colback=gray!5!white, 
    colframe=gray!60!black,
    title={
      \parbox[t]{\dimexpr\linewidth-4mm\relax}{%
        \ttfamily
        Task prompt for in silico antibody binding optimization
      }
    },
    breakable
  ]
  
Your task is to improve the properties of an antibody Fab by editing only Fv, while staying within a maximum number of de-gapped sequence edits from the starting molecule. \\

Properties are computed on the de-gapped Fv. Charge, GRAVY, liability, immunogenicity, edit\_distance, and sequence\_length use de-gapped heavy plus de-gapped light. AbLang2 naturalness is the mean of the separate de-gapped heavy-chain and light-chain scores.\\

SCORING

The objective properties are:

\{property\_description\}

  \end{tcolorbox}

\paragraph{In silico antibody binding optimization.} The task prompt is combined with the shared prompt.

  \begin{tcolorbox}[
    colback=gray!5!white, 
    colframe=gray!60!black,
    title={
      \parbox[t]{\dimexpr\linewidth-4mm\relax}{%
        \ttfamily
        Task prompt for in silico antibody binding optimization
      }
    },
    breakable
  ]
  
Your task is to improve binding of an antibody Fab to its target protein by editing only heavy-chain CDR-H3, while staying within a maximum number of de-gapped sequence edits from the starting molecule. \\

SCORING

The two objective properties are: \\
- pTM: Protenix global fold confidence (0–1). \\
- abag\_ipTM: antibody - antigen interface confidence (0–1), the primary binding signal.

  \end{tcolorbox}

\paragraph{Curator.} The insight curator gets the retrieved subgraph as input and generates the natural language insight.

  \begin{tcolorbox}[
    colback=gray!5!white, 
    colframe=gray!60!black,
    title={
      \parbox[t]{\dimexpr\linewidth-4mm\relax}{%
        \ttfamily
        Prompt for insight curator
      }
    },
    breakable
  ]
  You curate antibody-optimization memory insights for mutations within one germline. Return a JSON list with one object per input item, each containing exactly the string fields id and insight. Each insight must be 2-3 tightly written sentences covering the mechanism, when the mutation helps or hurts, and actionable guidance. The insight must add mechanistic, region-aware reasoning (framework vs CDR, biophysical rationale for the property) and actionable mutation guidance. Stay grounded in the supplied evidence. The region field, when present, is the authoritative AHo region (e.g. fr1, fr2, cdr2) for this position; state it verbatim and never infer the region from the position number. When amino\_acid\_groups is present, use the original-to-mutated residue R-group change for biophysical rationale.  \\

The item has two effect fields: single\_avg is the isolated single-mutation effect (only this mutation applied); avg is the mean full-design effect when this mutation appeared alongside others (confounded). Prefer single\_avg for mechanistic claims. The item may include co\_occurrence: each neighbor entry has single\_avg (neighbor's isolated effect), co\_avg (true isolated pair effect — only these two mutations applied), and epistasis\_avg (= co\_avg - this single\_avg- neighbor single\_avg), the deviation from additivity in the raw property: positive means co\_avg is greater than the sum of the singles, negative means less, near zero is additive. In your insight, when co\_occurrence is present: cite the neighbor token, give the numeric epistasis\_avg and its sign, and say whether it is favorable given the objective direction: positive for maximize, negative for minimize; for range, judge movement toward the target band. If the objective or target-band evidence is unavailable, describe only the deviation. Make combination claims ONLY from co\_occurrence; omit if absent.
  \end{tcolorbox}

\subsection{Hyperparameters}

\begin{table}[h]
\centering
\caption{\textbf{Hyperparameters used in all experiments.}}
\resizebox{0.7\linewidth}{!}{
\begin{tabular}{llc}
\toprule[1.25pt]
\textbf{Task} & \textbf{Hyperparameter} & \textbf{Value} \\
\midrule[1.25pt]
\multirow{3}{*}{Multiple property} & Rollouts per example & 10 \\
& Number of examples & 100 \\
& Maximum turns per rollout & 15 \\
\midrule
\multirow{3}{*}{In silico binding} & Rollouts per example & 30 \\
& Number of examples & \# complexes per cluster \\
& Maximum turns per rollout & 10 \\
\midrule
\multirow{3}{*}{Both} & Local memory top-$k$ & 60 \\
& Global memory top-$k$ & 30 \\
& \# of Neighbors $N$ & 6 \\
\bottomrule[1.25pt]
\end{tabular}}
\label{tab:hyperparameters}
\end{table}

For in silico antibody binding optimization, the number of examples per cluster is determined by the number of antigen--antibody complexes available in each cluster, as shown in \cref{tab:denovo_clusters}.

\newpage

\section{Additional experimental results}\label{appx:3_add_experiments}

\subsection{Cumulative performance progression without a rolling average}

\begin{figure}[h]
    \centering
    \includegraphics[width=\linewidth]{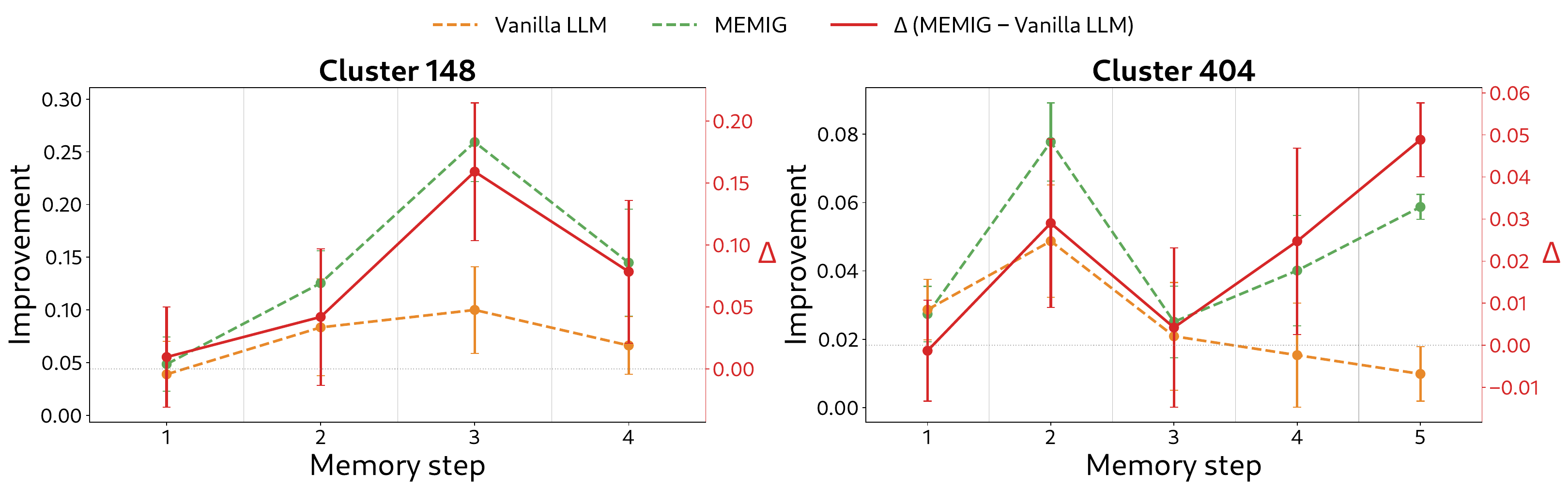}
    \caption{
        \textbf{Cumulative performance progression without a rolling average.} Vertical lines mark memory updates, where the rollouts between two lines are generated concurrently under the same memory. Each point is the mean improvement over all rollouts sharing a memory step, and error bars indicate the standard error across those rollouts. 
    }
    \label{fig_appx:cumulative_nonrolling}
    \vspace{-0.2in}
\end{figure}

The rolling average in \cref{fig:3_6_cumulative} smooths the rollouts within each memory step, which can obscure the value of an individual step. To remove this smoothing, we report the mean improvement of every rollout sharing a memory step as a single point in \cref{fig_appx:cumulative_nonrolling}, with error bars indicating the standard error across those rollouts. We observe the same trend as in \cref{fig:3_6_cumulative}, where \methodname improves over successive memory steps while the vanilla LLM does not, which confirms that the progression is not an artifact of the smoothing.

\subsection{Same oracle call analysis}\label{appx:same_oracle}

\begin{table}[h]
\centering
\caption{\textbf{Comparison under the same oracle calls.} \textbf{Bold} highlights the best score in a cluster and * denotes additional decomposition oracle call mode. }\label{tab:appx_same_oracle}
\small
\begin{tabular}{ccccc}
\toprule[1.25pt]
\textbf{Cluster} & \textbf{Model} & Pass@1 & Pass@3 & Avg \\
\midrule[1.25pt]
\multirow{3}{*}{148}
 & DC-Cu*          & 0.067 & 0.182 & 0.109 \\
 & ACE*           & 0.100 & 0.215 & \textbf{0.169} \\
 & \methodname (Ours)   & \textbf{0.258} & \textbf{0.456} & 0.144 \\
\midrule
\multirow{3}{*}{193}
 & DC-Cu*         & 0.056 & 0.159 & 0.110  \\
 & ACE*         & 0.028 & 0.081 & 0.107 \\
 & \methodname (Ours)   & \textbf{0.194} & \textbf{0.374} & \textbf{0.141} \\
\midrule
\multirow{3}{*}{993}
 & DC-Cu*          & 0.000 & 0.000 & 0.070 \\
 & ACE*            & 0.000 & 0.000 & 0.070 \\
 & \methodname (Ours)   & \textbf{0.007} & \textbf{0.020} & \textbf{0.090} \\
\bottomrule[1.25pt]
\end{tabular}
\end{table}

We compare \methodname against the memory baselines under a matched oracle budget, since the decomposition of \methodname consumes additional oracle calls. Here, single-mutation and pairwise decomposition is an epistasis graph component of \methodname rather than a shared experimental setting, since ACE~\citep{zhang2026ace} and DC-Cu~\citep{suzgun2026dynamic} accumulate experience from the trajectories an agent produces. Therefore, modifying them is not required for a fair comparison, as the decomposition is part of what we propose rather than an advantage granted by the setup. Nevertheless, we report a setting in which ACE and DC-Cu receive the same decomposed measurements, which removes any difference in the oracle call budget between the methods.

We provide the results in \cref{tab:appx_same_oracle} under GLM-5.3-flash backbone. We observe that \methodname attains the highest pass@1 and pass@3 on all three clusters even when the baselines share the same oracle calls, which shows that the gain comes from how the measurements are stored rather than from access to them. Here, ACE attains a higher average improvement on one cluster while satisfying the constraints less often, which is consistent with the observation that improvement is averaged over all designs regardless of whether they satisfy the constraint.

\subsection{Cost analysis} 

\begin{table*}[h]
\centering
\caption{\textbf{Token cost analysis.} Average token consumption per rollout. Input and output tokens are reported in thousands. Tokens per pass include tokens consumed by unsuccessful rollouts.}
\label{tab3_4_3:cost}
\small
\begin{tabular}{ccccc}
\toprule[1.25pt]
\textbf{Method} & \textbf{Input} & \textbf{Output} & \textbf{Pass@1} & \textbf{Tokens / pass} \\
\midrule[1.25pt]
\textbf{Vanilla LLM} & 98.1 & 6.8 & 0.022 & 4.719M \\
\textbf{Obs.} & 204.3 & 7.2 & 0.000 & -- \\
\textbf{ACE} & 103.8 & 6.6 & 0.017 & 6.623M \\
\textbf{DC-Cu} & 120.3 & 7.7 & 0.006 & 23.028M \\
\textbf{\methodname} & 270.1 & 9.2 & \textbf{0.194} & \textbf{1.436M} \\
\bottomrule[1.25pt]
\end{tabular}
\end{table*}

\begin{table}[h]
\centering
\caption{\textbf{Oracle evaluation cost.} Average number of oracle evaluations per rollout. Single and pair evaluations are additional evaluations used by \methodname to construct the epistasis graph.}
\label{tab:oracle_cost}
\small
\begin{tabular}{cccccc}
\toprule[1.25pt]
\textbf{Method} & \textbf{Ref.} & \textbf{Candidate} & \textbf{Single} & \textbf{Pair} & \textbf{Total} \\
\midrule[1.25pt]
\textbf{Vanilla LLM} & 1.00 & 6.03 & 0 & 0 & \phantom{0}7.03 \\
\textbf{Obs.} & 1.00 & 5.88 & 0 & 0 & \phantom{0}6.88 \\
\textbf{ACE} & 1.00 & 5.44 & 0 & 0 & \phantom{0}6.44 \\
\textbf{DC-Cu} & 1.00 & 5.76 & 0 & 0 & \phantom{0}6.76 \\
\textbf{\methodname} & 1.00 & 4.95 & 8.46 & 22.97 & 37.38 \\
\bottomrule[1.25pt]
\end{tabular}
\end{table}

We report the computational cost of GLM-5.3-Flash on the in silico antibody binding optimization task (cluster 193). As shown in \cref{tab3_4_3:cost}, \methodname uses more input tokens because the retrieved memory is injected before each rollout, while achieving the highest Pass@1 and the lowest token cost per successful rollout. While the absolute token usage is highest, the token cost per successful rollout reduces to 1.436M, compared with $>$4M for all baselines.

\cref{tab:oracle_cost} reports the mean number of unique oracle inputs per rollout. The additional single- and pair-mutation inputs introduced by \methodname's epistasis decomposition increase this for $\binom{n}{2}+n$ where n denotes the number of proposed mutations. This increase is by design, as \methodname explicitly evaluates isolated mutations and pairs to measure epistasis, and these additional evaluations remain practical in our in silico setting because the computational oracles are easily re-runnable without experimental measurements. Moreover, the decomposition budget is a tunable knob rather than a structural requirement. For example, pairs can be subsampled instead of exhaustively decomposed, so the design does not depend on oracles being free.

\subsection{Detailed results}

\paragraph{Multi property optimization.} We provide the detailed results for \cref{sec:3_2_multi}.

\begin{table*}[h]
\centering
\caption{\textbf{Multiple property optimization.} \textcolor{teal}{Teal} indicates improvement over all baselines and \textbf{bold} indicates the best results.}\label{tab:3_2_mpo}
\resizebox{0.8\linewidth}{!}{
\begin{tabular}{cc cc ccccc}
\toprule[1.25pt]
& & \multicolumn{2}{c}{\textbf{Pass}} & \multicolumn{5}{c}{\textbf{Improvement}} \\
\cmidrule(lr){3-4} \cmidrule(lr){5-9}
\textbf{Backbone} & \textbf{Model} & @1 & @3 & Avg. & Charge & GRAVY & Immu. & Liab. \\
\midrule[1.25pt]
\multicolumn{2}{c}{Random} & 0.003 & 0.009 & -0.419 & -0.071 & 0.000 & -1.243 & -0.363 \\
\multicolumn{2}{c}{MCMC} & 0.004 & 0.012 & -0.043 & 0.308 & 0.008 & -0.465 & -0.115 \\
\midrule[1.25pt]
\multirow{5}{*}{Haiku-4.5}
& Vanilla & 0.077 & 0.212 & 0.431 & 1.223 & 0.004 & -0.037 & 0.535 \\
& Obs. & 0.073 & 0.168 & 0.477 & 1.225 & 0.019 & 0.167 & 0.499 \\
& DC-Cu & 0.188 & 0.423 & 0.602 & 1.240 & 0.040 & 0.431 & 0.696 \\
& ACE & 0.036 & 0.094 & 0.390 & 1.255 & 0.032 & 0.197 & 0.078 \\
& \methodname (Ours)
& {\textcolor{teal}{0.221}}
& {{0.373}}
& {\textcolor{teal}{0.718}}
& 1.246
& 0.039
& {\textcolor{teal}{1.036}}
& 0.551 \\
\midrule[1.25pt]
\multirow{5}{*}{GLM-5.3-Flash}
& Vanilla & 0.424 & 0.699 & 1.327 & 1.273 & 0.044 & 0.287 & 3.708 \\
& Obs. & 0.461 & 0.717 & 1.344 & 1.256 & 0.044 & 0.560 & 3.515 \\
& DC-Cu & 0.409 & 0.707 & 1.621 & 1.262 & 0.044 & -0.165 & \textbf{5.343} \\
& ACE & 0.395 & 0.650 & 1.577 & 1.270 & 0.044 & -0.049 & 5.043 \\
& \methodname (Ours)
& \textbf{\textcolor{teal}{0.761}}
& \textbf{\textcolor{teal}{0.847}}
& \textbf{\textcolor{teal}{1.754}}
& \textbf{\textcolor{teal}{1.279}}
& \textbf{\textcolor{teal}{0.045}}
& \textbf{\textcolor{teal}{2.276}}
& 3.416 \\
\bottomrule[1.25pt]
\end{tabular}}
\end{table*}

\paragraph{In silico antibody binding optimization.}
We provide the detailed results for \cref{sec:3_3_insilico}.
\begin{table*}[h]
    \centering
    \caption{\textbf{In silico antibody binindg optimization.} \textbf{Bold} indicates the best per column across all methods and \textcolor{teal}{teal} indicates improvement over all non-structural methods within the same backbone. * indicates structural baselines included as reference.}\label{tab:3_3_denovo}

    \resizebox{\linewidth}{!}{
    \begin{tabular}{ll ccccc ccccc ccccc}
    \toprule[1.25pt]
    & & \multicolumn{5}{c}{993} & \multicolumn{5}{c}{114} & \multicolumn{5}{c}{147} \\
        \cmidrule(lr){3-7} \cmidrule(lr){8-12} \cmidrule(lr){13-17}
    & & \multicolumn{2}{c}{Pass} & \multicolumn{3}{c}{Improvement} & \multicolumn{2}{c}{Pass} & \multicolumn{3}{c}{Improvement} & \multicolumn{2}{c}{Pass} & \multicolumn{3}{c}{Improvement} \\
        \cmidrule(lr){3-4} \cmidrule(lr){5-7} \cmidrule(lr){8-9} \cmidrule(lr){10-12} \cmidrule(lr){13-14} \cmidrule(lr){15-17}
    \textbf{Category} & \textbf{Model} & @1 & @3 & Avg & ipTM & pTM & @1 & @3 & Avg & ipTM & pTM & @1 & @3 & Avg & ipTM & pTM \\
        \midrule[1.25pt]
    \multirow{2}{*}{Statistical}
        & Random  & 0.000 & 0.000 & 0.028 & 0.043 & 0.013 & 0.089 & 0.206 & 0.016 & 0.036 & -0.003 & 0.024 & 0.070 & -0.035 & -0.041 & -0.030 \\
        & MCMC    & \textbf{0.013} & \textbf{0.040} & 0.075 & 0.110 & 0.039 & 0.233 & 0.368 & 0.069 & 0.115 & 0.024 & 0.171 & 0.383 & 0.064 & 0.101 & 0.027 \\
        \midrule
    \multirow{2}{*}{Structure-based}
        & ProteinMPNN*  & 0.000 & 0.000 & 0.138 & \textbf{0.196} & 0.081 & 0.000 & 0.000 & 0.000 & 0.000 & 0.000 & 0.000 & 0.000 & 0.000 & 0.000 & 0.000 \\
        & AbMPNN*  & 0.000 & 0.000 & 0.116 & 0.164 & 0.068 & \textbf{0.600} & \textbf{0.762} & \textbf{0.160} & \textbf{0.242} & \textbf{0.077} & 0.229 & 0.406 & 0.109 & 0.158 & 0.059 \\
        \midrule[1.25pt]
    \multirow{5}{*}{Haiku-4.5}
        & Vanilla         & 0.000 & 0.000 & 0.010 & 0.014 & 0.007 & 0.122 & 0.164 & 0.053 & 0.081 & 0.025 & 0.048 & 0.128 & 0.044 & 0.065 & 0.024 \\
        & + Obs.          & 0.000 & 0.000 & 0.077 & 0.105 & 0.050 & 0.411 & 0.626 & 0.133 & 0.204 & 0.062 & 0.229 & 0.431 & 0.097 & 0.141 & 0.053 \\
        & + DC-Cu         & 0.000 & 0.000 & 0.054 & 0.076 & 0.032 & 0.306 & 0.366 & 0.098 & 0.151 & 0.045 & 0.162 & 0.347 & 0.099 & 0.145 & 0.053 \\
        & + ACE           & 0.000 & 0.000 & 0.048 & 0.068 & 0.028 & 0.283 & 0.333 & 0.098 & 0.149 & 0.046 & 0.052 & 0.129 & 0.090 & 0.129 & 0.050 \\
        & + \methodname (Ours) & 0.000 & 0.000 & \textbf{\textcolor{teal}{0.141}} & \textcolor{teal}{0.193} & \textbf{\textcolor{teal}{0.089}} & 0.317 & 0.397 & 0.130 & 0.197 & 0.062 & \textbf{\textcolor{teal}{0.533}} & \textbf{\textcolor{teal}{0.876}} & \textbf{\textcolor{teal}{0.217}} & \textbf{\textcolor{teal}{0.324}} & \textbf{\textcolor{teal}{0.109}} \\
        \midrule[1.25pt]
    \multirow{5}{*}{GLM-5.3-Flash}
        & Vanilla         & 0.000 & 0.000 & 0.042 & 0.056 & 0.028 & 0.156 & 0.291 & 0.053 & 0.082 & 0.024 & 0.067 & 0.159 & 0.063 & 0.090 & 0.035 \\
        & + Obs.          & 0.000 & 0.000 & 0.061 & 0.083 & 0.039 & 0.217 & 0.316 & 0.044 & 0.068 & 0.019 & 0.076 & 0.159 & 0.017 & 0.026 & 0.007 \\
        & + DC-Cu         & 0.000 & 0.000 & 0.042 & 0.058 & 0.026 & 0.178 & 0.313 & 0.033 & 0.051 & 0.015 & 0.052 & 0.124 & 0.028 & 0.041 & 0.016 \\
        & + ACE           & 0.000 & 0.000 & 0.048 & 0.064 & 0.032 & 0.256 & 0.365 & 0.091 & 0.141 & 0.041 & 0.114 & 0.252 & 0.036 & 0.055 & 0.017 \\
        & + \methodname (Ours) & \textcolor{teal}{0.007} & \textcolor{teal}{0.020} & \textcolor{teal}{0.090} & \textcolor{teal}{0.124} & \textcolor{teal}{0.056} & \textcolor{teal}{0.278} & 0.348 & \textcolor{teal}{0.099} & \textcolor{teal}{0.155} & \textcolor{teal}{0.043} & \textcolor{teal}{0.269} & \textcolor{teal}{0.480} & \textcolor{teal}{0.128} & \textcolor{teal}{0.191} & \textcolor{teal}{0.066} \\
     \bottomrule[1.25pt]
    \end{tabular}}

    \vspace{0.5em}

    \resizebox{\linewidth}{!}{
    \begin{tabular}{ll ccccc ccccc ccccc}
    \toprule[1.25pt]
    & & \multicolumn{5}{c}{148} & \multicolumn{5}{c}{193} & \multicolumn{5}{c}{234} \\
        \cmidrule(lr){3-7} \cmidrule(lr){8-12} \cmidrule(lr){13-17}
    & & \multicolumn{2}{c}{Pass} & \multicolumn{3}{c}{Improvement} & \multicolumn{2}{c}{Pass} & \multicolumn{3}{c}{Improvement} & \multicolumn{2}{c}{Pass} & \multicolumn{3}{c}{Improvement} \\
        \cmidrule(lr){3-4} \cmidrule(lr){5-7} \cmidrule(lr){8-9} \cmidrule(lr){10-12} \cmidrule(lr){13-14} \cmidrule(lr){15-17}
    \textbf{Category} & \textbf{Model} & @1 & @3 & Avg & ipTM & pTM & @1 & @3 & Avg & ipTM & pTM & @1 & @3 & Avg & ipTM & pTM \\
        \midrule[1.25pt]
    \multirow{2}{*}{Statistical}
        & Random  & 0.008 & 0.025 & -0.020 & -0.026 & -0.014 & 0.011 & 0.032 & 0.041 & 0.055 & 0.027 & 0.192 & 0.453 & -0.115 & -0.174 & -0.055 \\
        & MCMC    & 0.025 & 0.075 & 0.027 & 0.045 & 0.009 & 0.044 & 0.128 & 0.104 & 0.141 & 0.067 & 0.600 & 0.890 & 0.011 & 0.012 & 0.009 \\
        \midrule
    \multirow{2}{*}{Structure-based}
        & ProteinMPNN*  & 0.000 & 0.000 & 0.285 & 0.417 & 0.153 & 0.000 & 0.000 & 0.268 & 0.386 & 0.150 & 0.108 & 0.208 & \textbf{0.075} & \textbf{0.109} & \textbf{0.042} \\
        & AbMPNN*  & \textbf{0.575} & \textbf{0.866} & \textbf{0.288} & \textbf{0.421} & \textbf{0.155} & 0.000 & 0.000 & \textbf{0.285} & \textbf{0.401} & \textbf{0.169} & 0.733 & 0.775 & 0.052 & 0.075 & 0.030 \\
        \midrule[1.25pt]
    \multirow{5}{*}{Haiku-4.5}
        & Vanilla         & 0.075 & 0.203 & 0.051 & 0.075 & 0.026 & 0.017 & 0.049 & 0.031 & 0.043 & 0.018 & 0.600 & 0.920 & 0.037 & 0.054 & 0.020 \\
        & + Obs.          & 0.017 & 0.050 & 0.055 & 0.078 & 0.031 & 0.017 & 0.049 & 0.050 & 0.070 & 0.030 & 0.833 & \textbf{0.997} & 0.069 & 0.103 & 0.035 \\
        & + DC-Cu         & 0.067 & 0.182 & 0.091 & 0.132 & 0.050 & 0.033 & 0.098 & 0.168 & 0.229 & 0.107 & 0.958 & \textbf{1.000} & 0.067 & 0.100 & 0.034 \\
        & + ACE           & 0.192 & 0.335 & 0.132 & 0.193 & 0.072 & 0.117 & 0.262 & 0.183 & 0.253 & 0.112 & 0.967 & \textbf{1.000} & \textbf{0.075} & \textbf{0.109} & \textbf{0.042} \\
        & + \methodname (Ours) & \textcolor{teal}{0.342} & \textcolor{teal}{0.579} & \textcolor{teal}{0.201} & \textcolor{teal}{0.293} & \textcolor{teal}{0.109} & \textbf{\textcolor{teal}{0.139}} & \textbf{\textcolor{teal}{0.287}} & 0.171 & 0.237 & 0.105 & \textbf{\textcolor{teal}{0.975}} & \textbf{1.000} & 0.074 & 0.108 & 0.041 \\
        \midrule[1.25pt]
    \multirow{5}{*}{GLM-5.3-Flash}
        & Vanilla         & 0.025 & 0.075 & 0.072 & 0.103 & 0.041 & 0.022 & 0.067 & 0.085 & 0.114 & 0.055 & 0.575 & 0.799 & 0.015 & 0.023 & 0.007 \\
        & + Obs.          & 0.000 & 0.000 & 0.057 & 0.073 & 0.040 & 0.000 & 0.000 & 0.116 & 0.158 & 0.073 & 0.767 & 0.972 & 0.051 & 0.075 & 0.027 \\
        & + DC-Cu         & 0.017 & 0.050 & 0.025 & 0.036 & 0.014 & 0.006 & 0.017 & 0.044 & 0.059 & 0.029 & 0.792 & 0.930 & 0.035 & 0.053 & 0.018 \\
        & + ACE           & 0.083 & 0.214 & 0.036 & 0.050 & 0.022 & 0.017 & 0.049 & 0.108 & 0.146 & 0.070 & 0.675 & 0.917 & 0.053 & 0.072 & 0.034 \\
        & + \methodname (Ours) & \textcolor{teal}{0.258} & \textcolor{teal}{0.456} & \textcolor{teal}{0.144} & \textcolor{teal}{0.079} & \textcolor{teal}{0.210} & \textbf{\textcolor{teal}{0.194}} & \textbf{\textcolor{teal}{0.374}} & \textcolor{teal}{0.141} & \textcolor{teal}{0.197} & \textcolor{teal}{0.084} & \textcolor{teal}{0.925} & \textcolor{teal}{\textbf{1.000}} & \textcolor{teal}{0.067} & \textcolor{teal}{0.101} & 0.033 \\
     \bottomrule[1.25pt]
    \end{tabular}}

    \vspace{0.5em}

    \resizebox{\linewidth}{!}{
    \begin{tabular}{ll ccccc ccccc ccccc}
    \toprule[1.25pt]
    & & \multicolumn{5}{c}{268} & \multicolumn{5}{c}{308} & \multicolumn{5}{c}{348} \\
        \cmidrule(lr){3-7} \cmidrule(lr){8-12} \cmidrule(lr){13-17}
    & & \multicolumn{2}{c}{Pass} & \multicolumn{3}{c}{Improvement} & \multicolumn{2}{c}{Pass} & \multicolumn{3}{c}{Improvement} & \multicolumn{2}{c}{Pass} & \multicolumn{3}{c}{Improvement} \\
        \cmidrule(lr){3-4} \cmidrule(lr){5-7} \cmidrule(lr){8-9} \cmidrule(lr){10-12} \cmidrule(lr){13-14} \cmidrule(lr){15-17}
    \textbf{Category} & \textbf{Model} & @1 & @3 & Avg & ipTM & pTM & @1 & @3 & Avg & ipTM & pTM & @1 & @3 & Avg & ipTM & pTM \\
        \midrule[1.25pt]
    \multirow{2}{*}{Statistical}
        & Random  & 0.122 & 0.300 & -0.037 & -0.047 & -0.026 & 0.095 & 0.263 & 0.007 & 0.012 & 0.001 & 0.192 & 0.455 & -0.101 & -0.144 & -0.058 \\
        & MCMC    & 0.338 & 0.584 & 0.057 & 0.095 & 0.019 & 0.214 & 0.519 & 0.122 & 0.174 & 0.070 & 0.546 & 0.752 & 0.008 & 0.012 & 0.004 \\
        \midrule
    \multirow{2}{*}{Structure-based}
        & ProteinMPNN*  & 0.000 & 0.000 & 0.115 & 0.191 & 0.038 & 0.000 & 0.000 & 0.170 & 0.235 & 0.104 & 0.000 & 0.000 & -0.022 & -0.041 & -0.002 \\
        & AbMPNN*  & 0.074 & 0.130 & 0.117 & 0.198 & 0.037 & 0.014 & 0.042 & 0.059 & 0.081 & 0.038 & 0.237 & 0.328 & 0.057 & 0.081 & 0.033 \\
        \midrule[1.25pt]
    \multirow{5}{*}{Haiku-4.5}
        & Vanilla         & 0.351 & 0.568 & 0.081 & 0.130 & 0.032 & 0.195 & 0.337 & 0.051 & 0.073 & 0.029 & 0.421 & 0.635 & 0.025 & 0.037 & 0.013 \\
        & + Obs.          & 0.633 & 0.786 & 0.102 & 0.168 & 0.035 & 0.500 & 0.653 & 0.202 & 0.287 & 0.116 & 0.650 & 0.806 & 0.054 & 0.079 & 0.030 \\
        & + DC-Cu         & 0.163 & 0.332 & 0.015 & 0.024 & 0.006 & 0.386 & 0.711 & 0.179 & 0.255 & 0.103 & \textbf{0.842} & 0.875 & 0.048 & 0.071 & 0.026 \\
        & + ACE           & 0.653 & 0.748 & 0.088 & 0.146 & 0.030 & 0.238 & 0.515 & 0.137 & 0.193 & 0.081 & 0.796 & \textbf{0.887} & 0.054 & 0.078 & 0.029 \\
        & + \methodname (Ours) & \textbf{\textcolor{teal}{0.694}} & \textcolor{teal}{0.844} & \textbf{\textcolor{teal}{0.121}} & \textbf{\textcolor{teal}{0.199}} & \textbf{\textcolor{teal}{0.042}} & \textbf{\textcolor{teal}{0.729}} & \textbf{\textcolor{teal}{0.962}} & 0.171 & 0.246 & 0.096 & 0.838 & \textbf{0.887} & \textbf{\textcolor{teal}{0.097}} & \textbf{\textcolor{teal}{0.148}} & \textbf{\textcolor{teal}{0.046}} \\
        \midrule[1.25pt]
    \multirow{5}{*}{GLM-5.3-Flash}
        & Vanilla         & 0.307 & 0.547 & 0.040 & 0.068 & 0.012 & 0.233 & 0.541 & 0.148 & 0.210 & 0.086 & 0.575 & 0.764 & 0.021 & 0.033 & 0.009 \\
        & + Obs.          & 0.518 & 0.757 & 0.090 & 0.148 & 0.032 & 0.295 & 0.574 & 0.157 & 0.225 & 0.089 & 0.583 & 0.725 & -0.012 & -0.020 & -0.005 \\
        & + DC-Cu         & 0.194 & 0.386 & 0.033 & 0.055 & 0.012 & 0.295 & 0.638 & 0.158 & 0.225 & 0.092 & 0.492 & 0.599 & 0.000 & -0.001 & 0.002 \\
        & + ACE           & 0.318 & 0.524 & 0.038 & 0.064 & 0.012 & 0.248 & 0.569 & 0.113 & 0.160 & 0.065 & 0.683 & 0.817 & 0.017 & 0.029 & 0.006 \\
        & + \methodname (Ours) & \textcolor{teal}{0.692} & \textbf{\textcolor{teal}{0.877}} & \textcolor{teal}{0.103} & \textcolor{teal}{0.170} & \textcolor{teal}{0.036} & \textcolor{teal}{0.643} & \textcolor{teal}{0.913} & \textbf{\textcolor{teal}{0.228}} & \textbf{\textcolor{teal}{0.327}} & \textbf{\textcolor{teal}{0.128}} & \textcolor{teal}{0.804} & \textcolor{teal}{0.884} & \textcolor{teal}{0.070} & \textcolor{teal}{0.103} & \textcolor{teal}{0.036} \\
     \bottomrule[1.25pt]
    \end{tabular}}

    \vspace{0.5em}

    \resizebox{0.72\linewidth}{!}{
    \begin{tabular}{ll ccccc ccccc}
    \toprule[1.25pt]
    & & \multicolumn{5}{c}{404} & \multicolumn{5}{c}{409} \\
        \cmidrule(lr){3-7} \cmidrule(lr){8-12}
    & & \multicolumn{2}{c}{Pass} & \multicolumn{3}{c}{Improvement} & \multicolumn{2}{c}{Pass} & \multicolumn{3}{c}{Improvement} \\
        \cmidrule(lr){3-4} \cmidrule(lr){5-7} \cmidrule(lr){8-9} \cmidrule(lr){10-12}
    \textbf{Category} & \textbf{Model} & @1 & @3 & Avg & ipTM & pTM & @1 & @3 & Avg & ipTM & pTM \\
        \midrule[1.25pt]
    \multirow{2}{*}{Statistical}
        & Random  & 0.067 & 0.181 & -0.021 & -0.039 & -0.004 & 0.322 & 0.690 & 0.002 & -0.002 & 0.007 \\
        & MCMC    & 0.140 & 0.357 & 0.029 & 0.058 & 0.000 & 0.567 & 0.922 & 0.040 & 0.066 & 0.013 \\
        \midrule
    \multirow{2}{*}{Structure-based}
        & ProteinMPNN*  & 0.007 & 0.020 & 0.058 & 0.117 & 0.000 & 0.244 & 0.388 & 0.079 & 0.138 & 0.019 \\
        & AbMPNN*  & 0.293 & 0.517 & \textbf{0.070} & \textbf{0.141} & 0.000 & \textbf{0.989} & \textbf{1.000} & \textbf{0.086} & \textbf{0.149} & \textbf{0.023} \\
        \midrule[1.25pt]
    \multirow{5}{*}{Haiku-4.5}
        & Vanilla         & 0.187 & 0.443 & 0.045 & 0.090 & 0.000 & 0.656 & 0.917 & 0.061 & 0.105 & 0.017 \\
        & + Obs.          & 0.313 & 0.472 & 0.062 & 0.124 & 0.000 & 0.844 & 0.995 & 0.081 & 0.140 & \textbf{0.023} \\
        & + DC-Cu         & 0.133 & 0.356 & 0.051 & 0.103 & 0.000 & 0.778 & 0.906 & 0.033 & 0.057 & 0.010 \\
        & + ACE           & 0.280 & 0.614 & 0.064 & 0.128 & 0.000 & 0.978 & \textbf{1.000} & 0.074 & 0.131 & 0.017 \\
        & + \methodname (Ours) & \textbf{\textcolor{teal}{0.440}} & \textbf{\textcolor{teal}{0.744}} & 0.063 & 0.127 & 0.000 & 0.933 & 0.998 & 0.072 & 0.124 & 0.019 \\
        \midrule[1.25pt]
    \multirow{5}{*}{GLM-5.3-Flash}
        & Vanilla         & 0.047 & 0.131 & 0.025 & 0.050 & 0.000 & 0.522 & 0.816 & 0.033 & 0.055 & 0.011 \\
        & + Obs.          & 0.107 & 0.279 & 0.034 & 0.069 & -0.001 & 0.667 & 0.890 & 0.051 & 0.083 & 0.018 \\
        & + DC-Cu         & 0.107 & 0.279 & 0.022 & 0.045 & -0.001 & 0.611 & 0.729 & 0.024 & 0.037 & 0.011 \\
        & + ACE           & 0.100 & 0.263 & 0.040 & 0.081 & 0.000 & 0.711 & 0.890 & 0.045 & 0.074 & 0.015 \\
        & + \methodname (Ours) & \textcolor{teal}{0.260} & \textcolor{teal}{0.438} & \textcolor{teal}{0.062} & \textcolor{teal}{0.125} & 0.000 & \textcolor{teal}{0.956} & \textbf{\textcolor{teal}{1.000}} & \textcolor{teal}{0.077} & \textcolor{teal}{0.136} & \textcolor{teal}{0.019} \\
     \bottomrule[1.25pt]
    \end{tabular}}
\end{table*}

\end{document}